\pdfoutput=1

\documentclass[11pt]{article}
\usepackage[table]{xcolor}
\usepackage[final]{acl}

\usepackage{times}
\usepackage{latexsym}
\usepackage{makecell}
\usepackage{booktabs}
\usepackage{algorithm}
\usepackage{algpseudocode}
\usepackage{mathtools}
\usepackage{amssymb}
\usepackage{hyperref}
\usepackage[T1]{fontenc}
\usepackage{pifont}

\definecolor{lightgray}{gray}{0.92}
\definecolor{lightblue}{rgb}{0.8,0.85,1}
\definecolor{red2}{HTML}{F5B7B1}
\definecolor{green2}{HTML}{ABEBC6}
\usepackage[utf8]{inputenc}
\usepackage{tabularx}
\newcommand{\cmark}{\ding{51}}
\newcommand{\xmark}{\ding{55}}
\newcommand{\notcheckmark}{{\cmark}\textsuperscript{\textcolor{black}{\kern-0.7em{\bf---}}}}

\usepackage{microtype}
\usepackage{multirow}

\usepackage{inconsolata}
\usepackage{array}

\usepackage[textsize=scriptsize]{todonotes}
\usepackage{inconsolata}

\usepackage[skins,breakable]{tcolorbox}

\definecolor{pink-color}{RGB}{237,46,104} 
\definecolor{dark-grey-color}{RGB}{79,91,102}

\newcommand{\paramnorm}[1]{
\textcolor{pink-color}{\small{\texttt{\detokenize{#1}}}}
}

\newtcolorbox[list inside=prompt,auto counter,number within=section]{prompt}[1][]{
    colbacktitle=black!80,
    colframe=black!80,
    coltitle=white,
    fontupper=\footnotesize,
    boxsep=5pt,
    left=0pt,
    right=0pt,
    top=0pt,
    bottom=0pt,
    boxrule=1pt,
    enhanced, 
    breakable,
    skin first=enhanced,
    skin middle=enhanced,
    skin last=enhanced,
    #1,
}

\title{Learning to Refine with Fine-Grained Natural Language Feedback}

\author{Manya Wadhwa \quad\quad Xinyu Zhao  \quad\quad Junyi Jessy Li \quad\quad Greg Durrett \\
The University of Texas at Austin \\
    \texttt{manya.wadhwa@utexas.edu}
}

\begin{document}
\maketitle
\begin{abstract}
Recent work has explored the capability of large language models (LLMs) to identify and correct errors in LLM-generated responses. 
These refinement approaches frequently evaluate what sizes of models are able to do refinement for what problems, but less attention is paid to what effective feedback for refinement looks like. In this work, we propose looking at refinement with feedback as a composition of three distinct LLM competencies: (1) \textbf{detection} of bad generations; (2) fine-grained natural language \textbf{critique} generation; (3) \textbf{refining} with fine-grained feedback. The first step can be implemented with a high-performing discriminative model and steps 2 and 3 can be implemented either via prompted or fine-tuned LLMs. A key property of the proposed \textsc{Detect}, \textsc{Critique}, \textsc{Refine} (``DCR'') method is that the step 2 critique model can give fine-grained feedback about errors, made possible by offloading the discrimination to a separate model in step 1. We show that models of different capabilities benefit from refining with DCR on the task of improving factual consistency of document grounded summaries. Overall, DCR consistently outperforms existing end-to-end refinement approaches and current trained models not fine-tuned for factuality critiquing.\footnote{Code and models available at: \url{https://github.com/ManyaWadhwa/DCR}} 
\end{abstract}
\section{Introduction}

\begin{figure}[t!]
\centering
\includegraphics[scale=0.5, trim= 0 22cm 0 0]{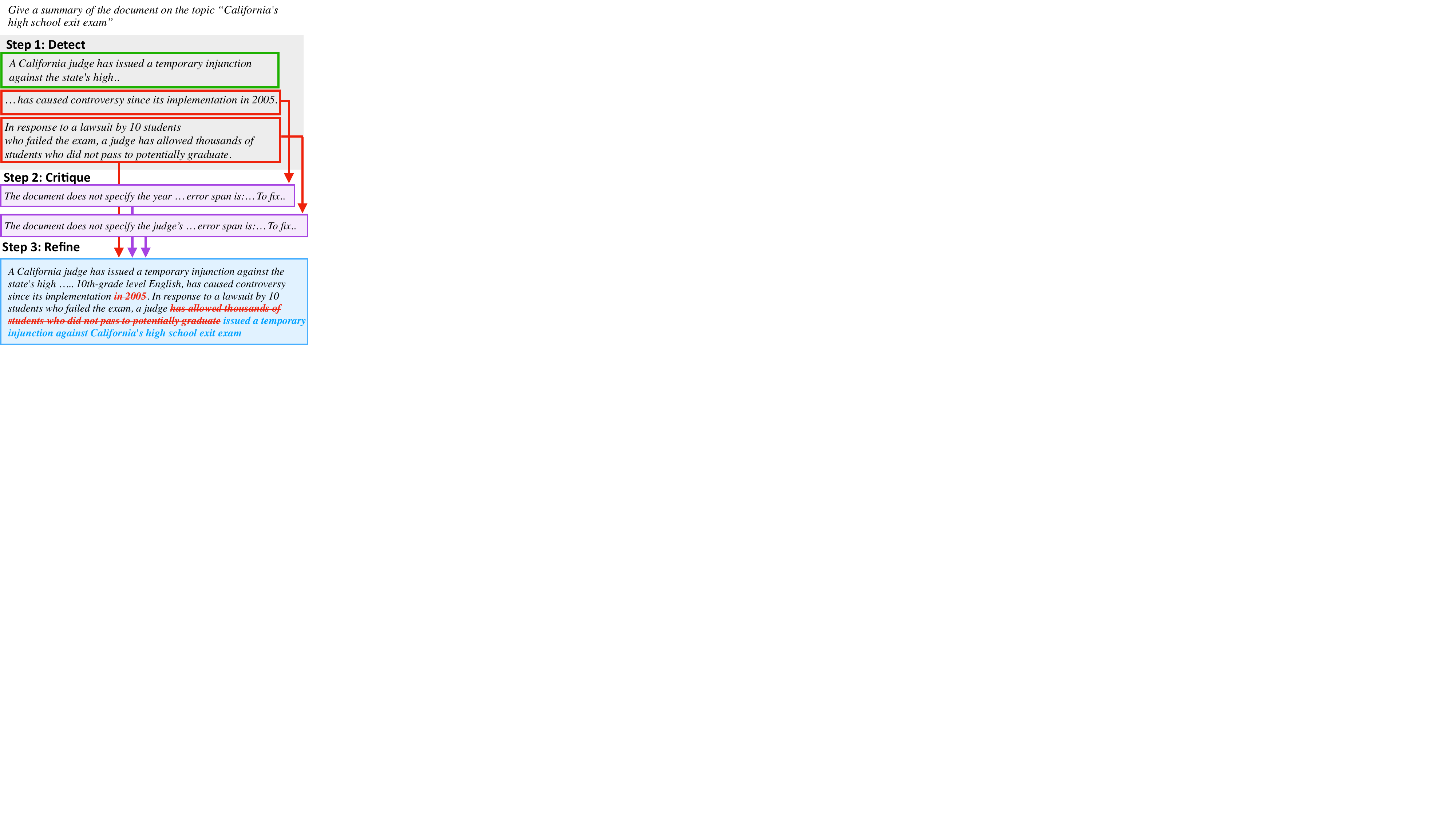}
\caption{ Overview of the proposed \textsc{Detect}, \textsc{Critique}, \textsc{Refine} method. For a document-grounded response, DCR first \textbf{detects} if the initial response should be refined. If so, we \textbf{critique} the response by generating fine-grained natural language feedback about errors, then \textbf{refine} by making targeted edits.}
\vspace{-1em}
\label{fig:motivation_figure}
\end{figure}

Large language models (LLMs) have been observed to display inconsistent behavior such hallucinations, not following instructions, and unfaithful reasoning \cite{levy-etal-2021-investigating,Ye-Durrett:2022:Fewshot,zhang2024how,turpin2024language,shaikh2023second, zhuo2023red} One recent strategy to fix these mistakes is to perform post-hoc refinement of the response with natural language feedback \cite{pan2024automatically,madaan2024self}. These methods either use human feedback \cite{saunders-2023-improving} or, more frequently, automated feedback, such as from self-critiquing \cite{madaan2024self, gero2023selfverification, shinn2024reflexion, raunak-etal-2023-leveraging, selfee2023}, from a trained model \cite{xu-etal-2024-llmrefine, akyurek-etal-2023-rl4f, paul-etal-2024-refiner, chern2024factool}, or from external tools \cite{jiang2023selfevolve, olausson2024is, gou2023critic,chen2024teaching, stengel2024regal}. 

Critiques are straightforward to obtain in some of these settings: for example, verifying that an acronym starts with the correct letters \cite{madaan2024self} or that source code passes test cases \cite{olausson2024is}. However, in the context of natural language generation, as opposed to code generation or math problem solving, notions of correctness are relatively less well defined. As a result, substantial prior work has taken a ``one-size-fits-all'' approach and either directly refined responses with an LLM, or generated feedback across a wide variety of aspects and then used those for refinement \cite{wang2023shepherd,selfee2023}. 
Prior work has not studied what happens when refining for a dimension like factual correctness, which has the property of not being easily verifiable with external tools while still being objective.

In this work, we propose a three-stage refinement framework suitable for tasks like factual correctness. We look at refinement as a composition of three distinct LLM competencies: \textsc{Detect}, \textsc{Critique} and \textsc{Refine} (DCR). Figure \ref{fig:motivation_figure} shows an example of this pipeline. We first detect erroneous generations at a sentence level. Then, if any
sentence is identified to have an error, we proceed to generate fine-grained natural language feedback describing the errors and how to fix them. Finally, we 
refine the original outputs
with the generated feedback. We show that models of different abilities perform better when refining with our proposed decomposition compared to baselines where the response is either (a) refined with a general instruction (e.g. ``\emph{improve factual consistency}'') \cite{saunders2022self} , or (b) refined with natural language feedback where the model needs to do verification in the feedback step \cite{madaan2024self}. This process allows for two key differences from prior work: first, the use of a detector to focus the feedback, and second, the ability to fine-tune models on fine-grained feedback to enumerate specific errors. This sentence-level approaches allows for individually enumerating many errors across an entire LLM output by handling them in a factored way. 

We evaluate our approach on two datasets of document-grounded LLM outputs: TofuEval \cite{Tang2024} and a subset of UltraChat \cite{ding-etal-2023-enhancing} consisting of queries asking for summaries. We believe these are representative tasks for a wider range of such use cases. Across both tasks, we show that our three-stage approach outperforms ablations removing or simplifying these stages. In addition, the form of feedback given by our models leads to higher factual consistency post-refinement than feedback from Shepherd \cite{wang2023shepherd} or SelFee \cite{selfee2023}. Finally, we show that fine-tuning our critique model improves its capabilities over prompting, and our model is able to give feedback on a variety of factual inconsistencies.

Our main contributions are: (1) we introduce a novel post-hoc refinement method: \textsc{Detect}, \textsc{Critique} and \textsc{Refine} (DCR), that refines with natural language feedback to enhance factual consistency; (2) we fine-tune models to generate fine-grained factual inconsistency localization, reasoning about the error, and a suggested fix for the inconsistency; (3) we show the importance of the \textsc{Detect} and \textsc{Critique} steps in enhancing the post-hoc refinement capabilities of models.

\begin{table*}[ht!]
\centering
\small
\renewcommand{\tabcolsep}{1.4mm}
\begin{tabular}{ll|ccccc}
\toprule
 & \makecell{Size} & \makecell{Factuality} & Document grounded & ``No Error'' cases & Error localization  & Eval.~~on refinement\\
\midrule
UltraCM & 13B & \cellcolor{red2}\xmark & \cellcolor{green2}\cmark & \cellcolor{red2}\xmark & \cellcolor{red2}\xmark  & \cellcolor{red2}\xmark \\
Shepherd & 13B & \cellcolor{red2}\xmark & \cellcolor{red2}\xmark & \cellcolor{red2}\xmark & \cellcolor{red2}\xmark  & \cellcolor{green2}\cmark \\
SelFee & 7B/13B & \cellcolor{red2}\xmark & \cellcolor{green2}\cmark & \cellcolor{green2}\cmark & \cellcolor{red2}\xmark & \cellcolor{green2}\cmark \\
\midrule
\makecell{DCR (Ours)} & 7B & \cellcolor{green2}\cmark & \cellcolor{green2}\cmark & \cellcolor{green2}\cmark  & \cellcolor{green2}\cmark & \cellcolor{green2}\cmark \\
\bottomrule
\end{tabular}
\vspace{-0.5em}
\caption{Comparison between existing feedback models and our trained model. Our approach focuses on generating fine-grained feedback for improving factual consistency of document-grounded responses.}
\label{tab:feedback_models}
\vspace{-1em}
\end{table*}

\section{Background and Task Setup}\label{sec:task_setup}

We assume we are given an LLM output $r$, generated from a document $D$ by prompting a model $M$; this accommodates tasks like traditional summarization, query-focused summarization, document-grounded question answering, and more. Our goal is to generate a refinement $\hat{r}=M'(r)$ where $M'$ is the refinement model, which can be distinct from $M$ in our setting. 
We have two conditions for refinement to be successful. First, we want to improve the quality of the response along the desired refinement axis, which in our case is factual consistency. We define a function $E(D,r)$ to score responses. For the tasks we consider, there is not a firm binary notion of factual consistency; this follows from work in NLI showing that entailment judgments are inherently subjective \cite{pavlick-kwiatkowski-2019-inherent,nie-etal-2020-learn,chen-etal-2020-uncertain}).
Therefore, we will evaluate if $E(D, \hat{r}) > E(D, r)$; that is, did our refinement successfully improve factual consistency. Second, we want to edit the response such the refinement preserves the style, structure and most of the content from the original response. We do not want the refinement process to simply replace the original response or delete large portions of it. We evaluate this by quantifying the number of edits at a word level and doing a qualitative analysis, but our first priority is to optimize for $E$.

\vspace{-0.5em}
\paragraph{Relation to past formalizations} 

Prior work like Self-Refine \cite{madaan2024self} leverages the source model to critique and refine its own output, which assumes that the source model has the capacity to follow a prompt and evaluate its own generations. These methods evaluate their refinement methods on more structured tasks with automatic metrics, such as solve rate for math reasoning \cite{Cobbe2021TrainingVT} and fraction of programs optimized for code optimization \cite{shypula2024learning}. Our work does not constrain the refinement model to be the same as the source, which changes the nature of the questions we investigate, and furthermore the factual consistency task has different properties than logical reasoning problems like math.

In domains like program synthesis, refinement is often compared with sampling more completions from the original model \cite{olausson2024is}. In this work, we assume as part of the problem definition that we are refining a base response $r$. This task is useful when generating the base response may be expensive, or if it may follow other constraints or instructions that make it challenging to regenerate. Furthermore, this allows us to use past datasets that annotate errors over responses \cite{Tang2024}, which enables us to perform more fine-grained analyses of fixed and remaining errors.

\vspace{-0.5em}
\paragraph{Prior critiquing methods} Table \ref{tab:feedback_models} shows a comparison of our proposed critique model with prior approaches like UltraCM, Shepherd and Selfee. Feedback from these models mostly focus on critiquing the overall quality of the response without necessarily verifying whether or not the respose needs refinement. While UltraCM does not evaluate the effectiveness of its feedback via refinement, Shepherd and SelFee evaluate on tasks like multiple choice QA \cite{wang-etal-2024-large-language-models-fair, Mihaylov2018CanAS, Lin2021TruthfulQAMH} where the evaluation objective is well-defined. Furthermore, current refinement methods often perform the \textsc{Critique} step directly on a response without knowing whether or not it \textit{needs} to be refined. This approach places the burden of both verification and critiquing on the same model, which our pipeline improves upon. 
\section{Refining with Fine-Grained Feedback}
We propose decomposing the task of refining textual responses using natural language feedback into three steps: \textsc{Detect}, \textsc{Critique}, \textsc{Refine}.  Algorithm \ref{alg:proposed_alg} concretely shows the cascade of these steps. 

\begin{algorithm}[t!]
\small
\caption{Detect, Critique, Refine}\label{alg:cap}
 \textbf{Input:}\\
 Document: $D$, Initial Response: $r$, Models {$M_{\text{detect}}$, $M_{\text{critique}}$, $M_{\text{refine}}$}\\
 \textbf{Output} Refined response $\hat{r}$
\begin{algorithmic}[1]
\State $\mathbf{s}$ = $\texttt{split}$($r$)
\State $F \leftarrow \varnothing$

\For {$s_i \in \mathbf{s}$}
 \If {$M_{\text{detect}}$($s_i$,$r$,  $D$)} \Comment{Detect} 
     \State $f_i = M_{\text{critique}} (s_i, r ,  D)$ \Comment{Critique}
     \State $F \leftarrow F   \cup f_i $
\EndIf
\EndFor
\State $\hat{r}=M_{\text{refine}} (F, r,  D)$ \Comment{Refine}
\State \textbf{return} $\hat{r}$
\end{algorithmic}
\label{alg:proposed_alg}
\end{algorithm}

\paragraph{Step 1: \textsc{Detect} with $M_{\text{detect}}$} For a response $r$ grounded in document $D$, we split the response into sentences
$\mathbf{s}=$\texttt{split}$(r)$ using NLTK, and for each sentence $s_i \in \mathbf{s}$, we determine if there is an error by computing $M_{\text{detect}}(s_i, r, D) \in \{0,1\}$. If all sentences are correct, we do not modify the response. If any sentence is marked with an error, we generate feedback using the \textsc{Critique} step.

\paragraph{Step 2: \textsc{Critique} with $M_{\text{critique}}$} Once we have determined $r$ to have an error, for each sentence $s_i$ with $M_{\text{detect}} = 1$, we generate a natural language feedback $f_i = M_{\text{critique}}(D,r, s_i)$ that does span localization, reasons about why the span has an error and then suggests a natural language fix
for the span. We combine the sentence wise feedback to create $F$ and use it for the \textsc{Refine} step. 

\paragraph{Step 3: \textsc{Refine} with $M_{\text{refine}}$} Given the document $D$, the response with errors $r$ and the natural language feedback $F$, we use $\hat{r} = M_{\text{refine}}(D,r,F)$ to generate a response which targets editing error spans mentioned in $F$. We refine with combined feedback as opposed to per-sentence to enable $M_\text{refine}$ to place edits in context and better produce a final coherent response. 

\subsection{Supervised Fine-Tuning} \label{sec:fine_tuning}
Section \ref{sec:results} shows how our proposed approach generalizes for models of different capabilities, ranging from \textsc{Llama-2-7b-Chat} to GPT-4. For smaller models, we explore fine-tuning $M_{\text{critique}}$ and $M_{\text{refine}}$ to generate natural language outputs for our task at hand. We train based on distilled critiques and refinements from a stronger language model $M_{\text{teacher}}$, in our case \textsc{gpt-4-0613}.

\begin{figure}[t!]
\centering
\includegraphics[trim={0 26cm 15cm 0},scale=0.35]{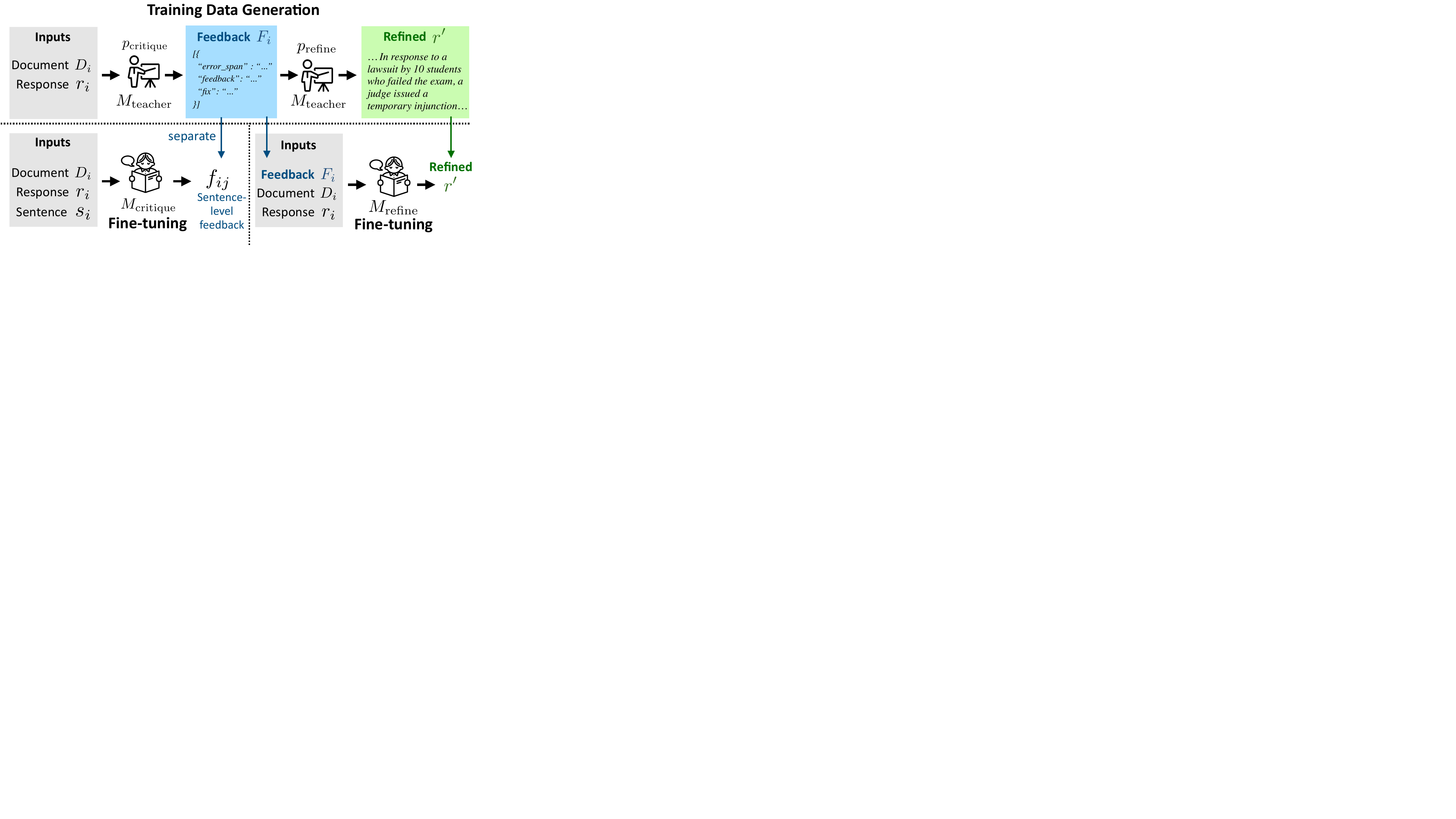}
\vspace{-2.5em}
\caption{Training data generation pipeline for our proposed models. We first generate structured feedback from $M_\text{teacher}$, convert it to a natural language form ($F$), and use that to generate a refinement ($r'$).}
\vspace{-1em}
\label{fig:data_generation_fig}
\end{figure}

Figure \ref{fig:data_generation_fig} gives an overview of the data generation and fine-tuning process. 
We first generate natural language feedback and refinements from $M_{\text{teacher}}$ using prompts $p_{\text{critique}}$ and $p_{\text{refine}}$. We then fine-tune $M_{\text{critique}}$ and $M_{\text{refine}}$. Notably, we first generate a \emph{structured} feedback to induce specific aspects from $M_{\text{teacher}}$ that would allow us to obtain training data for different capabilities.

\paragraph{Training Data Creation} We construct our fine-tuning data over a collection of (document, response) pairs $\{(D_i, r_i)\}_{i=1}^N$. We discard any responses for which $M_{\text{detect}}(s_i) = 0$ for all $s_i \in r$
such that $M_{\text{critique}}$ and $M_{\text{refine}}$ are only trained on responses containing errors.
For each $r_i$ detected to have an error, we prompt a teacher model $M_{\text{teacher}}$ using $p_{\text{critique}}$ to give a structured feedback. This is a list of objects where each object has an error span, reasoning as to why the span is an error, and a suggested fix. We convert this structure to a natural language form $F'_i$. The feedback prompt $p_{\text{critique}}$ anchors the feedback in error categories derived from prior work \cite{Tang2024}.

We then prompt $M_{\text{teacher}}$ using $p_{\text{refine}}$ to generate a refinement ${r'}_i$ using $F'_i$. As per our task setup, $p_{\text{refine}}$ contains a minimal editing instruction that guides the teacher model to make targeted edits. Prompts $p_{\text{critique}}$ and $p_{\text{refine}}$ for training data generation are given in Appendix \ref{append:ft}.

\paragraph{Fine Tuning} 
For $M_{\text{critique}}$, we use document $D_i$, response $r_i$ and a sentence from the response $s_{ij}$ as input. The model is then optimized to generated a natural language feedback $f_{ij}'$ for sentence $s_{ij}$. Note, that we train the model to output a feedback which has the error span, reasoning for the span being factually inconsistent, and the suggested fix.

We also fine-tune $M_{\text{refine}}$ in a similar manner. The input to this model is the document $D_i$, response $r_i$ and response-level feedback $F'_i$. The model is optimized to generate a refinement $r'_i$. Note that the $p_\text{critique}$ for data generation and fine-tuning are different. The prompts for fine-tuning are in Appendix \ref{subsec:baselines}. Appendix \ref{append:ft} describes the compute and the hyperparameters for fine-tuning.

\section{Experimental Setup}

\subsection{Datasets} \label{sec:datasets}

We consider two datasets for our task of post-hoc refinement: UltraChat \cite{ding-etal-2023-enhancing} and  MediaSum \cite{zhu-etal-2021-mediasum}, with annotations from TofuEval \cite{Tang2024}. In both datasets, we focus on refining document-grounded summaries to improve their factual consistency. With UltraChat, we create a train/val/test set by sampling summarization instructions from the dataset. For TofuEval, we use the MediaSum split as the test set and sample from the original train set of MediaSum to create the train/val set for our task. This gives us a distinct set of summaries from those in the TofuEval dataset (derived from MediaSum's test set).  Appendix \ref{appen:data_creation} discusses the data creation process in more detail.

\begin{table}[t!]
\centering
\small
\renewcommand{\tabcolsep}{1mm}
\renewcommand{\arraystretch}{1.1}
\begin{tabular}{llcccc}
\toprule
Subset & Dataset & \makecell{Size} & Doc Len & Resp Len & \% Correct\\
\midrule
Train & MediaSum & 1344 & 1189 & 43 & 46.9\\
 & UltraChat & 1072 & 486 & 225 &50.0\\
\midrule
Val & MediaSum & 149 & 1186 & 45 & 47.0\\
& UltraChat & 124 &524 & 238 & 50.0\\
\midrule
Test & TofuEval &267& 778 & 52 &56.6\\
& UltraChat &272& 497 & 227 &24.2\\
\bottomrule
\end{tabular}
\vspace{-0.5em}
\caption{Statistics of training/validation/test split for MediaSum/Tofueval and Ultrachat. 
}
\label{tab:synthetic_data}
\end{table}

\paragraph{Initial Response Generation} We run UltraChat and MediaSum prompts (source documents + summarization instructions) through \textsc{llama2-7b-chat} \cite{touvron2023llama} and generate initial responses. To be consistent with TofuEval's summary generation process, we ensure that MediaSum summaries are restricted to be less than 50 words (prompt \ref{prompt:initial_response_mediasum}). We combine the train/val splits from UltraChat and MediaSum for fine-tuning. For evaluation we use the MediaSum split of TofuEval, and the responses generated from Llama2 for UltraChat.  Appendix \ref{subsec:initial_response} shows examples of instructions and initial responses. Table \ref{tab:synthetic_data} shows the number of responses in each set along with their statistics. 

\begin{table*}[t!]
\centering
\small
\renewcommand{\tabcolsep}{0.6mm}
\rowcolors{3}{white}{lightgray} 
\begin{tabular}{l|lll|cccccc||cccccc}
\toprule
\multirow{2}{*}{\textbf{Strategy}}&
\multirow{2}{*}{\textbf{$M_\text{detect}$}} & 
\multirow{2}{*}{\textbf{$M_\text{critique}$}}& 
\multirow{2}{*}{\textbf{$M_\text{refine}$}} & 
\multicolumn{6}{c}{MediaSum/TofuEval} & 
\multicolumn{6}{c}{UltraChat} \\

&&&& \multicolumn{1}{c}{$\Delta A\uparrow$}  & 
\multicolumn{1}{c}{$\Delta G\uparrow$} & 
$W\uparrow$ & $S$ & $L$ & $\Delta MCS\uparrow$  & 
\multicolumn{1}{c}{$\Delta A\uparrow$ } & 
\multicolumn{1}{c}{$\Delta G\uparrow$ } & 
$W\uparrow$ & $S$ & $L$ &  $\Delta MCS\uparrow$\\ 
\midrule
DR  &  - & - & \makecell{Refine-\\L3-FT} & 0.05$^\dagger$ & 0.20$^\dagger$ &0.11$^\dagger$ & 0.88 & 0.01& - & 0.02$^\dagger$ & \textbf{0.08} &0.06$^\dagger$ & 0.94 & 0.0 & - \\

\makecell[l]{Feed + \\Refine} & - & \makecell{Critique-\\L3-FT} & \makecell{Refine-\\L3-FT} &  0.01$^\dagger$ & 0.04$^\dagger$ &0.01$^\dagger$ & 0.97 & 0.01 & - & 0.0$^\dagger$ & 0.02& 0.00$^\dagger$ & 1 & 0.0 & - \\

\makecell[l]{\textsc{Detect}+\\DR} & \makecell{Mini\\Check} & - & \makecell{Refine-\\L3-FT} &  0.05$^\dagger$ & 0.19$^\dagger$ & 0.09$^\dagger$ & 0.89 & 0.01 & 7.49$^\dagger$ & 0.00$^\dagger$  &\textbf{ 0.08} &0.03$^\dagger$ & 0.94 & 0.0 & 4.41$^\dagger$ \\

\makecell[l]{\textsc{DCR}\\(proposed)} & \makecell{Mini\\Check} & \makecell{Critique-\\L3-FT} & \makecell{Refine-\\L3-FT} &   \textbf{0.08} &  \textbf{0.33} &\textbf{0.17}  &  0.82 & 0.01 & \textbf{22.10}  & \textbf{0.06}   & 0.04  &\textbf{0.14}  &0.82 & 0.04 & \textbf{23.90}  \\
\midrule
DR  & - & - & \makecell{Refine-\\L2-FT} &  0.01$^\dagger$ & 0.00$^\dagger$ & 0.01$^\dagger$ & 0.98& 0.0 & - & 0.00$^\dagger$  &  0.03 & 0.0$^\dagger$ & 1.0 & 0.0 & - \\

\makecell[l]{Feed + \\Refine} & - & \makecell{Critique-\\L2-FT} &  \makecell{Refine-\\L2-FT} & -0.01$^\dagger$ & -0.04$^\dagger$ & 0.0$^\dagger$ & 0.99 & 0.00 & - & 0.0$^\dagger$ &0.02 & 0.00$^\dagger$ & 1.00 & 0.00 & - \\

\makecell[l]{\textsc{Detect}+\\DR}  & \makecell{Mini\\Check} & - &  \makecell{Refine-\\L2-FT} &  0.04$^\dagger$ & \textbf{0.22} & 0.07$^\dagger$ &0.91&0.01 & 8.24$^\dagger$ & \textbf{0.01}  & 0.01  & 0.02$^\dagger$ & 0.96 & 0.0 & 5.88$^\dagger$\\

\makecell[l]{\textsc{DCR}\\(proposed)}  & \makecell{Mini\\Check} & \makecell{Critique-\\L2-FT} &  \makecell{Refine-\\L2-FT} & \textbf{0.09} &  0.15  & \textbf{0.13}  & 0.82 & 0.04 & \textbf{19.10}  & 0.03  & -0.10 & \textbf{0.07}  & 0.76 & 0.09 & \textbf{21.32} \\    
\midrule
DR &- & - & GPT-4 &  0.03$^\dagger$ & 0.22$^\dagger$ & 0.13$^\dagger$ &0.87 & 0.0  & - & 0.01$^\dagger$ & 0.08$^\dagger$ & 0.06$^\dagger$ & 0.94 & 0.0\\
\makecell[l]{Feed + \\Refine} & & GPT-4 & GPT-4 &  0.10  & 0.49 &\textbf{0.25}& 0.73 & 0.01 & - &0.09 & \textbf{0.28} &\textbf{0.21} & 0.74&0.04 & -\\
\makecell[l]{\textsc{Detect}+\\DR}  & \makecell{Mini\\Check} & - & GPT-4 & \textbf{0.11} & 0.47  &0.19& 0.81 & 0.0 & 17.23$^\dagger$   & 0.04$^\dagger$ &  0.27 & 0.19 & 0.81 & 0.0 &  5.51$^\dagger$ \\
\makecell[l]{\textsc{DCR}\\(proposed)} & \makecell{Mini\\Check} & GPT-4 & GPT-4 & 0.10 &  \textbf{0.53} & 0.21 & 0.78 &0.02 & \textbf{19.85} & \textbf{0.07} & 0.21 & 0.18 &0.80&0.02 &  \textbf{22.79} \\

\bottomrule
\end{tabular}
\caption{Downstream evaluation of refinements as generated by our proposed method \textbf{DCR} and various refinement strategies. $^\dagger$:Statistically significant gains from DCR over these methods with $p<0.05$ according to a paired bootstrap test. DCR generally achieves the strongest performance across all base LLMs compared to other approaches, particularly on \textsc{Llama2-7b-Chat} and \textsc{Llama3-8b-Instruct}, and particularly on MCS.}
\label{table:main_results}
\end{table*}

\subsection{Refinement Strategies: Baselines} \label{subsec:baselines}

\paragraph{One-Step: Direct Refinement (DR)} 
\cite{welleck2023generating, chen2022codet, saunders2022self}. We generate a refinement $\hat{r} = M_{\text{refine}}(r)$ by directly prompting (or fine-tuning) the model with a general refinement instruction to improve the factual consistency of the initial response. For fine-tuning $M_\text{refine}$ for this baseline, we train with a balanced set of factually consistent and inconsistent responses (to prevent any copying behavior). The model is optimized to generate a GPT-4 distilled refinement if the initial response is inconsistent or simply copy the response if it is consistent.

\paragraph{Two-Step: Refinement with Natural Language Feedback (Feed+Refine)} 
This approach follows prior work using a two-step version of our pipeline \cite{madaan2024self,saunders2022self,akyurek-etal-2023-rl4f}. We first prompt the model to generate a natural language feedback for all sentences in the initial response $F = \cup M_{\text{critique}}(s_i) \forall s_i \in r$. We then generate a refinement that is on the feedback $\hat{r} = M_{\text{refine}}(F,r)$. 

We train $M_\text{critique}$ with a balanced set of consistent and inconsistent sentences. The model is optimized to generate `no error' if the sentence is factually consistent and the GPT-4 distilled feedback if the sentence is factually inconsistent. Similarly, $M_\text{refine}$ is trained with a balanced dataset of consistent and inconsistent summaries, and the refinement is conditioned on the feedback.   $M_\text{refine}$ is optimized to give a GPT-4 distilled refinement if the initial response is inconsistent or simply copy the response if it is consistent. 

\paragraph{Two-Step: Direct Refinement with \textsc{Detect} (Detect+DR)} We modify DR by first determining if the response is factually inconsistent or not $l_{r} = \cup M_\text{detect}(s_i) \forall s_i \in r$. If the response is factually inconsistent, we perform refinement with DR $\hat{r} = M_{\text{refine}}(r)$ if $l_{r}=1$. Note, this baseline does not use any external or self-generation feedback. We fine-tune $M_\text{refine}$ for this baseline and train it to generate GPT-4 distilled refinements for inconsistent responses.

\paragraph{Three-Step:  \textsc{Detect}, \textsc{Critique} - \textsc{Refine}, DCR (our method)} We first filter initial responses that are factually consistent using $M_\text{detect}$. We use $M_\text{critique}$ to generate sentence-wise feedback for any sentences that were detected to have an error. We combine this feedback and use $M_\text{refine}$ to make targeted changes. Models for this baseline are trained as described in Section \ref{sec:fine_tuning}. 

Prompts associated with each of the baselines are listed in Appendix \ref{append:refinement_prompts}. Note, that all refinement models are prompted (and fine-tuned) with a minimum editing instruction.

\subsection{Models}\label{subsec:models}

\paragraph{\textsc{Detect}} We use MiniCheck \cite{tang2024minicheck} as $M_\text{detect}$.
MiniCheck performs on par with GPT-4 while being light weight and more cost-friendly. This model is ideal for our approach since it does sentence level verification of factual consistency against a source document.

\paragraph{\textsc{Critique} and \textsc{Refine}} In order to test the effectiveness of our proposed method, we experiment with models of different capabilities:
GPT-4, \textsc{Llama-3-8b-Instruct} \cite{metallama3} and \textsc{Llama-2-7b-chat} \cite{touvron2023llama}.  We abbreviate non-fine-tuned versions of Llama as L2 and L3. We fine-tune Llama models to serve as $M_\text{critique}$ and $M_\text{refine}$ separately. The resulting Llama 3 models are referred to as Critique-L3-FT for $M_\text{critique}$ and Refine-L3-FT for $M_\text{refine}$, and analogously for Llama 2. 

\paragraph{Existing models as $M_{\text{critique}}$} We evaluate how existing feedback models \textsc{Shepherd} \cite{wang2023shepherd}\footnote{Note that \citet{wang2023shepherd} did not open source their model weights, so we use the model from the community that has been trained on their data: \url{https://huggingface.co/reciprocate/shepherd-13b}}, \textsc{UltraCM} 
\cite{cui2023ultrafeedback} and \textsc{SelFee} \cite{selfee2023} perform as $M_\text{critique}$. We run end-to-end refinements, varying $M_\text{critique}$ with \textsc{GPT-4}, and non fine-tuned versions of \textsc{Llama-3-8b-Instruct} and \textsc{Llama-2-7b-chat} as $M_\text{refine}$.

\subsection{Evaluation Metrics} \label{subsec:eval}

\paragraph{AlignScore ($\Delta A$)} AlignScore \cite{zha-etal-2023-alignscore} scores two texts in terms of general ``information alignment'' on a scale from 0-1 using RoBERTa \cite{liu2019roberta} as the base model. We report the delta in AlignScore which is the difference AlignScore(document, refined response) $-$ AlignScore(document, initial response).

\paragraph{GPT-4 Factuality Likert Scale Score ($\Delta G$)} 
Leveraging GPT-4's ability to score generations when given a well-defined rubric \cite{li2024leveraging} we prompt \textsc{gpt-4-0613} to score the factual consistency of a generation on a scale of 1-5 using a rubric. We score the initial and the refined response in independent GPT-4 calls and report the delta between them. The scoring prompt with the rubric is given in Appendix \ref{appen:eval_gpt4}.

\paragraph{GPT-4 Win-Rate ($W$, $S$, $L$)} We run pairwise scoring of the initial response and refinement using \textsc{gpt-4-0613} \cite{chiang2024chatbot,dubois2024alpacafarm} and prompt it to score generations on a scale of 1-5. We use the scores to determine the win rate. In each call we randomize the order of the two responses. We report the fractions of Wins ($W$), Same scores ($S$) and Losses ($L$). The scoring prompt with the rubric is given in Appendix \ref{appen:eval_gpt4_pairwise}.

\begin{table*}[t!]
\centering
\small
\rowcolors{3}{white}{lightgray} 
\begin{tabular}{l|l|ccccc||ccccc}
\toprule
  \multirow{2}{*}{\textbf{$M_\text{critique}$}}& \multirow{2}{*}{\textbf{$M_\text{refine}$}} & \multicolumn{5}{c}{MediaSum/TofuEval} & \multicolumn{5}{c}{UltraChat} \\
&& \multicolumn{1}{c}{$\Delta A\uparrow$} & \multicolumn{1}{c}{$\Delta G\uparrow$} & $W\uparrow$ & $S$ & $L$ & \multicolumn{1}{c}{$\Delta A\uparrow$} & \multicolumn{1}{c}{$\Delta G\uparrow$} & $W \uparrow$ & $S$ & $L$\\ 
\midrule
Shepherd   & L3   & 0.05$^\dagger$	&0.09$^\dagger$ &0.06$^\dagger$ &	0.90	&0.04 & 0.05$^\dagger$	&0.00 &0.10$^\dagger$ &	0.85&	0.05\\
UltraCM  & L3 & 0.03$^\dagger$ &	0.09$^\dagger$ & 0.05$^\dagger$ &	0.90 &	0.05 & 0.05$^\dagger$	&0.04& 0.08$^\dagger$	&0.84&	0.07  \\
SelFee 7b  & L3   & 0.04 & -0.03 & 0.15 & 0.64 & 0.21 &	0.00 & 0.04 & 0.11 & 0.74 & 0.15 \\
SelFee 13b & L3   & 0.02 & -0.15 & 0.12 & 0.57 & 0.31 & 0.00 & 0.05 & 0.12 & 0.71 & 0.16 \\
L3     & L3    & 0.07  & 0.27   &0.16 & 	0.80	& 0.03 & 0.05  & 0.08  & 0.08$^\dagger$	&0.88&	0.04 \\
Critique-L3-FT     & L3    & \textbf{0.10}  & \textbf{0.39} & 0.16	&0.82&	0.02 & \textbf{0.08}  & \textbf{0.19}  & \textbf{0.18} &	0.79&	0.04 \\
Critique-L3-FT   & Refine-L3-FT  & 0.08  &  0.33  & \textbf{0.17}  &  0.82 & 0.01 & 0.06   & 0.04  &0.14  & 0.82 & 0.04  \\
\midrule 
Shepherd   & L2   &   -0.01$^\dagger$ &	-0.10$^\dagger$ & 0.03$^\dagger$ &	0.87&	0.09 &	-0.01$^\dagger$&	-0.03 &0.06	&0.83	&0.11 \\
UltraCM        & L2    &  -0.01$^\dagger$	&-0.13$^\dagger$	& 0.04$^\dagger$ &	0.84	&0.12&	-0.02$^\dagger$ &	\textbf{0.04} & 0.06	&0.88&	0.07 \\
SelFee 7b & L2 & 0.00 & -0.49 & 0.03 & 0.55 & 0.41 & -0.01 & -0.08 & 0.08 & 0.73 & 0.20 \\
SelFee 13b     & L2    & -0.02 & -0.54 & 0.05 & 0.55 & 0.40 & -0.01 & \textbf{0.04} & 0.07 & 0.79 & 0.14 \\
L2     & L2    & -0.03$^\dagger$ & -0.27$^\dagger$ &	0.03$^\dagger$ & 0.88 &	0.1 & -0.05$^\dagger$	 & -0.50$^\dagger$  &	0.02 &	0.65 & 0.32 \\
Critique-L2-FT     & L2    &  0.01$^\dagger$ & -0.27$^\dagger$ &0.08$^\dagger$	 & 0.75 &	0.17 & 0.01	 & -0.11  &	 \textbf{0.11} &	0.81	& 0.08\\
Critique-L2-FT & Refine-L2-FT  & \textbf{0.09} &  \textbf{0.15} & \textbf{0.13} & 0.82 & 0.04 & \textbf{0.03} & -0.10 & 0.07 & 0.76 & 0.09\\
\midrule 
Shepherd   & GPT-4  &  0.06$^\dagger$ & 0.32$^\dagger$ & 0.10$^\dagger$ & 0.88 & 	0.03 & 0.06$^\dagger$ &	0.26$^\dagger$ & 0.11$^\dagger$	&0.87	&0.02\\
UltraCM   & GPT-4   & 0.05 & 0.21  & 0.08$^\dagger$ &	0.91 &	0.01 &  0.03$^\dagger$ &	0.13  & 0.10$^\dagger$ &	0.88 &	0.02\\
SelFee 7b & GPT-4  & 0.04 & 0.17 & 0.20 &	0.71 &	0.9 & 0.02 & 0.44$^\dagger$ & 0.21 & 0.68 & 0.11 \\
SelFee 13b & GPT-4 & 0.03 & 0.24 & 0.19 &	0.71 &	0.10 & 0.01 & \textbf{0.40} & \textbf{0.19} & 0.71 &	0.10   \\
GPT-4  & GPT-4   &  \textbf{0.10} &  \textbf{0.53} & \textbf{0.21} & 0.78 &0.02 & \textbf{0.07}  & 0.21  & 0.18  &0.80&0.02   \\
\bottomrule
\end{tabular}
\caption{Results from DCR while varying the $M_\text{critique}$ to evaluate existing feedback models  on our task. $^\dagger$: significant gains by DCR with a $p<0.05$ according to a paired bootstrap test. We see the that our proposed $M_\text{critique}$ achieves the largest gains across all metrics when compared to refining with feedback from existing critic models.}
\label{table:other_feedback_models}
\end{table*}

\paragraph{MiniCheck score ($\Delta$\% MCS)} We calculate the difference in the percentage of factually correct summaries before and after refinement, as detected by $M_\text{detect}$ i.e., MiniCheck. For fairness, we only use this metric to compare among methods that use $M_\text{detect}$ as a part of the refinement process i.e. \textsc{Detect}+ DR and our proposed method DCR.

\section{Results} \label{sec:results}

\paragraph{Does the three-step refinement help over standard refinement strategies?} Table \ref{table:main_results} shows how our proposed method compares against existing refinement baselines using the metrics defined in Section \ref{subsec:eval}. DCR gives the largest improvement in $\Delta A$ when refining with (Critique-L3-FT, Refine-L3-FT) and (Critique-L2-FT, Refine-L2-FT).  We observe a similar trend for $W$. When refining with GPT-4, we observe DCR performing much better than DR, and being on par with Feed+Refine and \textsc{Detect}+DR. We attribute this to GPT-4 being a stronger model and achieving closer to ideal refinement already. DCR also leads to a larger improvement in the fraction of summaries improved ($\Delta MCS$) compared to \textsc{Detect}+DR. This points to the importance of refining with fine-grained feedback beyond the \textsc{Detect} step. Tables \ref{tab:absolute_results_mediasum_1} and \ref{tab:absolute_results_ultrachat_1} show the average values for AlignScore, GPT-4 Score and pairwise GPT-4 score for the original response and the refinement.

Table \ref{tab:refinement_examples} shows examples of refinements generated by GPT-4, Refine-L3-FT and Refine-L2-FT using DCR. The edits are localized to a sentence or phrase. The changes made are sophisticated and add the correct information instead of trivially deleting factually inconsistent information. Furthermore, even our smaller-scale models are making similar edits to GPT-4. 

\vspace{-0.5em}
\paragraph{How do existing feedback models refine compare to the proposed critic model?} Table \ref{table:other_feedback_models} shows the effectiveness of the DCR-generated feedback by comparing it against refining with feedback from existing critic models. Using DCR as the refinement strategy, we vary $M_\text{critique}$ and use GPT-4, \textsc{Llama3-8b-Instruct} and \textsc{Llama2-7b-Chat} as $M_\text{refine}$.

The proposed feedback leads to the highest gains in refinements across all our metrics, datasets and models. This is expected since the existing models were trained to give an ``overall'' summary level feedback instead of an aspect-specific fine-grained feedback, which our model learns to generate.
Also, refining with a fine-tuned $M_\text{critique}$ and non fine-tuned $M_\text{refine}$ gives larger improvements over refining with non fine-tuned $M_\text{critique}$ and $M_\text{refine}$. This shows that smaller models can be fine-tuned to give more effective feedback which is useful for refinement.
Table \ref{tab:feedback_model_output} shows examples of feedback generated by different models. The proposed fine-grained feedback does error localization and also suggests a fix, inheriting such structure from our structured prompts on stronger models. In contrast, feedback from existing models, focus more on the missing details rather than factual consistency even when prompted for the latter.

\section{Understanding generated feedback}\label{sec:nlf}

\paragraph{Is the proposed detailed feedback form helpful?} The feedback used in our proposed method has two important parts: error localization and a reasoning for why it is an error with a suggested fix. To evaluate the importance of this detailed feedback, we fine-tune $M_\text{critique}$ to generate only the error localization as feedback and refine with a non fine-tuned $M_\text{refine}$. Focusing on \textsc{Llama3-8b-Instruct}, we show in Table \ref{tab:level_of_detail} that refining with our proposed feedback form (rows 2 \& 3) does significantly better than refining with a less detailed feedback i.e. only error localization. We can particularly see this on $\Delta G$ and $W$. This points to the effectiveness of using a more detailed feedback for refinement and validates the usefulness of our proposed feedback.

\begin{table}[t!]
\centering
\small
\renewcommand{\tabcolsep}{0.5mm}
\rowcolors{3}{white}{lightgray} 
\begin{tabular}{l|ccc||ccc}
\toprule
Feedback Detail & \multicolumn{3}{c}{TofuEval} & \multicolumn{3}{c}{UltraChat} \\
Critique w/& $\Delta A\uparrow$ & $\Delta G\uparrow$ & $W\uparrow$ & $\Delta A\uparrow$ & $\Delta G\uparrow$ & $W\uparrow$\\
\midrule
 \makecell[l]{ localization} & 0.08 & 0.19$^\dagger$ & 0.13$^\dagger$ & 0.05 & 0.04 & 0.08$^\dagger$ \\
 
 \ \ +feedback & \textbf{0.10}  & \textbf{0.39}  & 0.16 & \textbf{0.08}  & \textbf{0.19}  & \textbf{0.18}  \\
 

 \ \ + FT $M_{\text{refine}}$ &0.08  &  0.33  & \textbf{0.17}  &  0.06   & 0.04  &0.14   \\
\bottomrule                        
\end{tabular}
\caption{
Comparison of refining with our proposed feedback form versus refining with just error localization as feedback. $^\dagger$:significant with $p<0.05$.}
\label{tab:level_of_detail}
\vspace{-1em}
\end{table}

\paragraph{How does the generated feedback compare against human-written feedback?}
The MediaSum split of TofuEval has human-written explanations of why a sentence in the initial response is factually inconsistent with the source document. We leverage these to calculate sentence-level recall statistics by comparing them against the feedback from Feed+Refine and DCR. We divide these in the following categories (1) ``Error Match'': when both the human and model generated feedback discuss the same error (2) ``Error, No Match'': when the human and model generated feedback discuss different errors (3) ``No Error Detected, No Match'': when the human written explanation talks about an error but the model generated feedback says no error. We prompt GPT-4 (\textsc{gpt-4-0613}) with Prompt \ref{prompt:gt_comparison} to evaluate the above.

In Table \ref{tab:gt_match_against_feedback} we see the feedback generated from Critique-L3-FT and Critique-L2-FT using DCR has a significantly higher match rate compared when to Feed+Refine. In the latter, we see the model only learns to say ``no error''. We also see how beneficial fine-tuning is for DCR when using \textsc{Llama3-8b-Instruct} (L2) as $M_\text{critique}$; however, \textsc{Llama3-8b-Instruct} (L3) benefits substantially less. Table~\ref{tab:comparison_to_groundtruth} shows examples of human annotation and Critique-L3-FT feedback on TofuEval.

\begin{table}[t!]
\centering
\small
\renewcommand{\tabcolsep}{0.7mm}
\rowcolors{2}{white}{lightgray} 
\begin{tabular}{l|l|ccc}
\toprule
\makecell{$M_\text{critique}$} & \makecell{Refinement\\Strategy} & \makecell{Error\\Match}$\uparrow$ & \makecell{Error\\No Match}$\downarrow$ & \makecell{No Error \\Detected\\ No Match}$\downarrow$\\
\midrule
\makecell[l]{CRTQ-L3-FT} & Feed+Refine & 0.01 & 0.0 & 0.99 \\
\makecell[l]{CRTQ-L3-FT} & DCR & 0.58  & 0.06 & 0.36 \\
L3 & DCR & 0.56 & 0.08 & 0.36 \\ 
\midrule
\makecell[l]{CRTQ-L2-FT} & Feed+Refine & 0.01  & 0.0 & 0.99 \\
\makecell[l] {CRTQ-L2-FT}  & DCR &  0.58  & 0.06 & 0.36 \\
L2 & DCR & 0.1 & 0.02 & 0.87 \\
\midrule
GPT-4 & Feed+Refine & 0.76 & 0.06 &  0.18  \\
GPT-4 & DCR & 0.61 & 0.01 & 0.38  \\
\bottomrule
\end{tabular}
\caption{Sentence-wise comparison of generated feedback against human-written feedback in TofuEval. CRTQ is prefixed for models trained as $M_\text{critique}$.}
\label{tab:gt_match_against_feedback}
\end{table}

\paragraph{What kind of edits does the feedback model suggest and the refinement make?} We 
manually examine 50 feedbacks generated from GPT-4, Critique-L3-FT, and Critique-L2-FT when prompted to refine with DCR. Table \ref{tab:edit_distribution} shows the distribution (in percentage) of suggested edits based on error span granularity and edit actions. We observe variation in granularity of error spans and the edit actions suggested across all models, with phrase substitution being the most common. This is further supported by the breakdown of edit types in Table \ref{tab:edits}. 

\paragraph{Are the edits made by the refinement model faithful to the feedback?} We manually examined 50 MediaSum/TofuEval refinements generated using DCR by GPT-4, Refine-L3-FT. and Refine-L2-FT. For each instance, we first look at the error span and the reasoning and then look for evidence in the refinement for whether or not the error span was fixed according to the feedback. If the error span was fixed as per the feedback, we mark the edit being faithful to the feedback. For each response, we calculate the percentage of feedback points that were incorporated in the refinement and then average that over the 50 examples for every model.
We find that on average GPT-4 is able to incorporate 92\% of the feedback, where as Refine-L3-FT is able to incorporate 96\% and Refine-L2-FT is able to incorporate 69\%. Table \ref{tab:feedback_refinement_examples} shows examples of the original response, feedback and refinements as generated by Refine-L3-FT when refining with DCR. We see that the feedback extracts error spans along with reasons for why the span was an error and a suggested fix. We also see the refinement incorporating these changes in the final generation.

\section{Related Work}

Several recent evaluation datasets in NLP have followed the trend of collecting explanations alongside evaluation scores. This ranges from model based evaluation datasets \cite{jiang2024tigerscore,xu-etal-2023-instructscore,li2024leveraging,kim2024evallm, cui2023ultrafeedback, kim2023prometheus,kim2024prometheus} to human evaluation \cite{Trienes:2024:ACL, wadhwa2024using, saunders2022self}. In our work, we test feedback for effectiveness at refinement, which feedback for evaluation is not always optimal for. 

For refinement specifically, \citet{liu-etal-2023-improving} implement a two step (\textsc{Feed+Refine}) refinement pipeline for the task of improving factual consistency in summarization.However, their use of the XSum dataset results in several key differences: the high prevalence of errors, simple summarization task, and short summaries mean that no ``detect'' step is necessary. Furthermore, our use of stronger LLMs and factuality evaluators substantially changes the performance regime of our base model and evaluator methods, leading to different conclusions. \citet{fatahi-bayat-etal-2023-fleek} implement a two-step (\textsc{Detect+DR}) pipeline for claim-level fact-checking. They use an external knowledge source to retrieve the relevant evidence for the claim. They verify using the evidence and also use it to guide the revision. In our experiments, we show that the proposed three-step pipeline outperforms two-step refinement. \citet{xu-etal-2024-llmrefine} also compare feedback modalities like in our work. However, they focus on refining with a general instruction vs.~using scalar feedback vs.~using a binary feedback. 
Furthermore, they improve overall quality rather than a specific aspect like factuality.

To refine outputs for factuality, \citet{mishra2024finegrained} train a hallucination detector and editor by fine-tuning the model to localize fine-grained hallucination error types by tagging spans and then removing the tagged spans in post-editing. 
\citet{balachandran-etal-2022-correcting,fabbri-etal-2022-improving, thorne-vlachos-2021-evidence} train post-editing models with techniques like infilling and sentence-compression to train better post-editing models. These approaches do not use natural language feedback, and are most optimized for deletion or replacement of simple errors rather than complex rewriting, where feedback can more clearly articulate a subtle error.

\section{Conclusion}

In this work, we propose a new post-hoc refinement method: \textsc{Detect}, \textsc{Critique} and \textsc{Refine} (DCR). We showed that our method performs better than existing refinement baselines on the task of improving factual consistency in document-grounded topic-focused summaries. We also showed that smaller models can be fine-tuned to perform fine-grained feedback generation for identifying and reasoning about any factual inconsistency. When refining with our proposed method, smaller models perform on par with GPT-4 on our task. We also show that our critic model produces more effective feedback for the task of post-hoc refining as compared to existing models. 

\section*{Limitations}

Our work is scoped to focus on refining LLM responses to improve factual consistency. Factuality has two important properties as an evaluation dimension: (1) it has a mostly objective notion of correctness (annotators can largely agree on what is hallucinated or not); (2) refining a response may involve many small corrections to different parts of a response. Other aspects of LLM responses such as completeness and stylistic consistency share these problem features; however, we focus on factuality due to the existence of models for automatic evaluation. Nevertheless, we believe our approach can be generalized to other evaluation dimensions as well, potentially leveraging new models such as Prometheus \cite{kim2024prometheus}.

Our work proposes additional steps over baseline methods, which increases the inference-time cost. However, given that reliable generations are important and the performance gains we observe for smaller models, we believe the tradeoff is worth the extra compute.

Our work also relies on an off-the-shelf and reliable \textsc{Detect} model for doing sentence-level factual consistency detection. We understand that such a discriminator might not always be available for different aspects and also that subjective tasks cannot be classified as correct and incorrect. More exploration needs to be done on how to effectively choose and train $M_\text{detect}$ for tasks other than document-grounded factuality detection. Furthermore, DCR currently only fixes errors that are detected in the \textsc{Detect} stage. Follow-up work could explore the ability of the generator to compensate for detector errors.

We also note that the fine-tuning data comes from a similar distribution of documents and instructions as the test data. It is also limited to English. It remains to be seen how our work extends to other languages and general document-grounded tasks. However, our approach is not fundamentally restricted to English-language refinement in these domains.


\section*{Acknowledgments}

This work was principally supported by a grant from Open Philanthropy, as well as NSF CAREER Awards IIS-2145280, IIS-2145479, and the NSF AI Institute for Foundations of Machine Learning (IFML). Thanks to Karim Villaescusa F., Kathryn Kazanas, and Keziah Reina for human annotations for the task of editing with feedback. Thanks to Fangcong Yin for help with debugging fine-tuning code.

\bibliography{anthology,custom}

\begin{thebibliography}{64}
\expandafter\ifx\csname natexlab\endcsname\relax\def\natexlab#1{#1}\fi

\bibitem[{Akyurek et~al.(2023)Akyurek, Akyurek, Kalyan, Clark, Wijaya, and Tandon}]{akyurek-etal-2023-rl4f}
Afra~Feyza Akyurek, Ekin Akyurek, Ashwin Kalyan, Peter Clark, Derry~Tanti Wijaya, and Niket Tandon. 2023.
\newblock \href {https://doi.org/10.18653/v1/2023.acl-long.427} {{RL}4{F}: Generating natural language feedback with reinforcement learning for repairing model outputs}.
\newblock In \emph{Proceedings of the 61st Annual Meeting of the Association for Computational Linguistics (Volume 1: Long Papers)}, pages 7716--7733, Toronto, Canada. Association for Computational Linguistics.

\bibitem[{Balachandran et~al.(2022)Balachandran, Hajishirzi, Cohen, and Tsvetkov}]{balachandran-etal-2022-correcting}
Vidhisha Balachandran, Hannaneh Hajishirzi, William Cohen, and Yulia Tsvetkov. 2022.
\newblock \href {https://doi.org/10.18653/v1/2022.emnlp-main.667} {Correcting diverse factual errors in abstractive summarization via post-editing and language model infilling}.
\newblock In \emph{Proceedings of the 2022 Conference on Empirical Methods in Natural Language Processing}, pages 9818--9830, Abu Dhabi, United Arab Emirates. Association for Computational Linguistics.

\bibitem[{Chen et~al.(2022)Chen, Zhang, Nguyen, Zan, Lin, Lou, and Chen}]{chen2022codet}
Bei Chen, Fengji Zhang, Anh Nguyen, Daoguang Zan, Zeqi Lin, Jian-Guang Lou, and Weizhu Chen. 2022.
\newblock {CodeT: Code Generation with Generated Tests}.
\newblock \emph{arXiv preprint arXiv:2207.10397}.

\bibitem[{Chen et~al.(2020)Chen, Jiang, Poliak, Sakaguchi, and Van~Durme}]{chen-etal-2020-uncertain}
Tongfei Chen, Zhengping Jiang, Adam Poliak, Keisuke Sakaguchi, and Benjamin Van~Durme. 2020.
\newblock \href {https://doi.org/10.18653/v1/2020.acl-main.774} {Uncertain natural language inference}.
\newblock In \emph{Proceedings of the 58th Annual Meeting of the Association for Computational Linguistics}, pages 8772--8779, Online. Association for Computational Linguistics.

\bibitem[{Chen et~al.(2024)Chen, Lin, Sch{\"a}rli, and Zhou}]{chen2024teaching}
Xinyun Chen, Maxwell Lin, Nathanael Sch{\"a}rli, and Denny Zhou. 2024.
\newblock \href {https://openreview.net/forum?id=KuPixIqPiq} {Teaching large language models to self-debug}.
\newblock In \emph{The Twelfth International Conference on Learning Representations}.

\bibitem[{Chern et~al.(2024)Chern, Chern, Chen, Yuan, Feng, Zhou, He, Neubig, and Liu}]{chern2024factool}
I-Chun Chern, Steffi Chern, Shiqi Chen, Weizhe Yuan, Kehua Feng, Chunting Zhou, Junxian He, Graham Neubig, and Pengfei Liu. 2024.
\newblock \href {https://openreview.net/forum?id=jolYuxpVn1} {{FacTool: Factuality Detection in Generative {AI} - A Tool Augmented Framework for Multi-Task and Multi-Domain Scenarios}}.

\bibitem[{Chiang et~al.(2024)Chiang, Zheng, Sheng, Angelopoulos, Li, Li, Zhang, Zhu, Jordan, Gonzalez et~al.}]{chiang2024chatbot}
Wei-Lin Chiang, Lianmin Zheng, Ying Sheng, Anastasios~Nikolas Angelopoulos, Tianle Li, Dacheng Li, Hao Zhang, Banghua Zhu, Michael Jordan, Joseph~E Gonzalez, et~al. 2024.
\newblock {Chatbot Arena: An Open Platform for Evaluating LLMs by Human Preference}.
\newblock \emph{arXiv preprint arXiv:2403.04132}.

\bibitem[{Cobbe et~al.(2021)Cobbe, Kosaraju, Bavarian, Chen, Jun, Kaiser, Plappert, Tworek, Hilton, Nakano, Hesse, and Schulman}]{Cobbe2021TrainingVT}
Karl Cobbe, Vineet Kosaraju, Mohammad Bavarian, Mark Chen, Heewoo Jun, Lukasz Kaiser, Matthias Plappert, Jerry Tworek, Jacob Hilton, Reiichiro Nakano, Christopher Hesse, and John Schulman. 2021.
\newblock \href {https://api.semanticscholar.org/CorpusID:239998651} {Training verifiers to solve math word problems}.
\newblock \emph{ArXiv}, abs/2110.14168.

\bibitem[{Cui et~al.(2023)Cui, Yuan, Ding, Yao, Zhu, Ni, Xie, Liu, and Sun}]{cui2023ultrafeedback}
Ganqu Cui, Lifan Yuan, Ning Ding, Guanming Yao, Wei Zhu, Yuan Ni, Guotong Xie, Zhiyuan Liu, and Maosong Sun. 2023.
\newblock {UltraFeedback: Boosting Language Models with High-quality Feedback}.
\newblock \emph{arXiv preprint arXiv:2310.01377}.

\bibitem[{Ding et~al.(2023)Ding, Chen, Xu, Qin, Hu, Liu, Sun, and Zhou}]{ding-etal-2023-enhancing}
Ning Ding, Yulin Chen, Bokai Xu, Yujia Qin, Shengding Hu, Zhiyuan Liu, Maosong Sun, and Bowen Zhou. 2023.
\newblock \href {https://doi.org/10.18653/v1/2023.emnlp-main.183} {Enhancing chat language models by scaling high-quality instructional conversations}.
\newblock In \emph{Proceedings of the 2023 Conference on Empirical Methods in Natural Language Processing}, pages 3029--3051, Singapore. Association for Computational Linguistics.

\bibitem[{Dubois et~al.(2024)Dubois, Li, Taori, Zhang, Gulrajani, Ba, Guestrin, Liang, and Hashimoto}]{dubois2024alpacafarm}
Yann Dubois, Chen~Xuechen Li, Rohan Taori, Tianyi Zhang, Ishaan Gulrajani, Jimmy Ba, Carlos Guestrin, Percy~S Liang, and Tatsunori~B Hashimoto. 2024.
\newblock {Alpacafarm: A simulation framework for methods that learn from human feedback}.
\newblock \emph{Advances in Neural Information Processing Systems}, 36.

\bibitem[{Fabbri et~al.(2022)Fabbri, Choubey, Vig, Wu, and Xiong}]{fabbri-etal-2022-improving}
Alex Fabbri, Prafulla~Kumar Choubey, Jesse Vig, Chien-Sheng Wu, and Caiming Xiong. 2022.
\newblock \href {https://doi.org/10.18653/v1/2022.emnlp-main.623} {Improving factual consistency in summarization with compression-based post-editing}.
\newblock In \emph{Proceedings of the 2022 Conference on Empirical Methods in Natural Language Processing}, pages 9149--9156, Abu Dhabi, United Arab Emirates. Association for Computational Linguistics.

\bibitem[{Fatahi~Bayat et~al.(2023)Fatahi~Bayat, Qian, Han, Sang, Belyy, Khorshidi, Wu, Ilyas, and Li}]{fatahi-bayat-etal-2023-fleek}
Farima Fatahi~Bayat, Kun Qian, Benjamin Han, Yisi Sang, Anton Belyy, Samira Khorshidi, Fei Wu, Ihab Ilyas, and Yunyao Li. 2023.
\newblock \href {https://doi.org/10.18653/v1/2023.emnlp-demo.10} {{FLEEK}: Factual error detection and correction with evidence retrieved from external knowledge}.
\newblock In \emph{Proceedings of the 2023 Conference on Empirical Methods in Natural Language Processing: System Demonstrations}, pages 124--130, Singapore. Association for Computational Linguistics.

\bibitem[{Gero et~al.(2023)Gero, Singh, Cheng, Naumann, Galley, Gao, and Poon}]{gero2023selfverification}
Zelalem Gero, Chandan Singh, Hao Cheng, Tristan Naumann, Michel Galley, Jianfeng Gao, and Hoifung Poon. 2023.
\newblock \href {https://openreview.net/forum?id=SBbJICrglS} {Self-verification improves few-shot clinical information extraction}.
\newblock In \emph{ICML 3rd Workshop on Interpretable Machine Learning in Healthcare (IMLH)}.

\bibitem[{Gou et~al.(2024)Gou, Shao, Gong, Shen, Yang, Duan, and Chen}]{gou2023critic}
Zhibin Gou, Zhihong Shao, Yeyun Gong, Yelong Shen, Yujiu Yang, Nan Duan, and Weizhu Chen. 2024.
\newblock {CRITIC: Large Language Models Can Self-Correct with Tool-Interactive Critiquing}.

\bibitem[{Hu et~al.(2022)Hu, yelong shen, Wallis, Allen-Zhu, Li, Wang, Wang, and Chen}]{hu2022lora}
Edward~J Hu, yelong shen, Phillip Wallis, Zeyuan Allen-Zhu, Yuanzhi Li, Shean Wang, Lu~Wang, and Weizhu Chen. 2022.
\newblock \href {https://openreview.net/forum?id=nZeVKeeFYf9} {Lo{RA}: Low-rank adaptation of large language models}.
\newblock In \emph{International Conference on Learning Representations}.

\bibitem[{Jiang et~al.(2024)Jiang, Li, Zhang, Huang, Lin, and Chen}]{jiang2024tigerscore}
Dongfu Jiang, Yishan Li, Ge~Zhang, Wenhao Huang, Bill~Yuchen Lin, and Wenhu Chen. 2024.
\newblock \href {https://openreview.net/forum?id=SIojR1ruNQ} {{TIGERScore: Building Explainable Metric for All Text Generation Task}}.

\bibitem[{Jiang et~al.(2023)Jiang, Wang, and Wang}]{jiang2023selfevolve}
Shuyang Jiang, Yuhao Wang, and Yu~Wang. 2023.
\newblock {SelfEvolve: A Code Evolution Framework via Large Language Models}.
\newblock \emph{arXiv preprint arXiv:2306.02907}.

\bibitem[{Kim et~al.(2023)Kim, Shin, Cho, Jang, Longpre, Lee, Yun, Shin, Kim, Thorne et~al.}]{kim2023prometheus}
Seungone Kim, Jamin Shin, Yejin Cho, Joel Jang, Shayne Longpre, Hwaran Lee, Sangdoo Yun, Seongjin Shin, Sungdong Kim, James Thorne, et~al. 2023.
\newblock {Prometheus: Inducing evaluation capability in language models}.
\newblock In \emph{NeurIPS 2023 Workshop on Instruction Tuning and Instruction Following}.

\bibitem[{Kim et~al.(2024{\natexlab{a}})Kim, Suk, Longpre, Lin, Shin, Welleck, Neubig, Lee, Lee, and Seo}]{kim2024prometheus}
Seungone Kim, Juyoung Suk, Shayne Longpre, Bill~Yuchen Lin, Jamin Shin, Sean Welleck, Graham Neubig, Moontae Lee, Kyungjae Lee, and Minjoon Seo. 2024{\natexlab{a}}.
\newblock {Prometheus 2: An Open Source Language Model Specialized in Evaluating Other Language Models}.
\newblock \emph{arXiv preprint arXiv:2405.01535}.

\bibitem[{Kim et~al.(2024{\natexlab{b}})Kim, Lee, Shin, Kim, and Kim}]{kim2024evallm}
Tae~Soo Kim, Yoonjoo Lee, Jamin Shin, Young-Ho Kim, and Juho Kim. 2024{\natexlab{b}}.
\newblock {EvalLM: Interactive Evaluation of Large Language Model Prompts on User-Defined Criteria}.
\newblock In \emph{Proceedings of the CHI Conference on Human Factors in Computing Systems}, pages 1--21.

\bibitem[{{Leandro von Werra, Younes Belkada, Lewis Tunstall, Edward Beeching, Tristan Thrush, Nathan Lambert, and Shengyi Huang}(2020)}]{trl}
{Leandro von Werra, Younes Belkada, Lewis Tunstall, Edward Beeching, Tristan Thrush, Nathan Lambert, and Shengyi Huang}. 2020.
\newblock \href {{https://github.com/huggingface/trl}} {{TRL: Transformer Reinforcement Learning}}.

\bibitem[{Levenshtein et~al.(1966)}]{levenshtein1966binary}
Vladimir~I Levenshtein et~al. 1966.
\newblock {Binary codes capable of correcting deletions, insertions, and reversals}.
\newblock In \emph{Soviet physics doklady}, volume~10, pages 707--710. Soviet Union.

\bibitem[{Levy et~al.(2021)Levy, Saxon, and Wang}]{levy-etal-2021-investigating}
Sharon Levy, Michael Saxon, and William~Yang Wang. 2021.
\newblock \href {https://doi.org/10.18653/v1/2021.findings-acl.416} {Investigating memorization of conspiracy theories in text generation}.
\newblock In \emph{Findings of the Association for Computational Linguistics: ACL-IJCNLP 2021}, pages 4718--4729, Online. Association for Computational Linguistics.

\bibitem[{Li et~al.(2024)Li, Xu, Shen, Xu, Gu, and Tao}]{li2024leveraging}
Zhen Li, Xiaohan Xu, Tao Shen, Can Xu, Jia-Chen Gu, and Chongyang Tao. 2024.
\newblock {Leveraging Large Language Models for NLG Evaluation: A Survey}.
\newblock \emph{arXiv preprint arXiv:2401.07103}.

\bibitem[{Lin et~al.(2021)Lin, Hilton, and Evans}]{Lin2021TruthfulQAMH}
Stephanie~C. Lin, Jacob Hilton, and Owain Evans. 2021.
\newblock \href {https://api.semanticscholar.org/CorpusID:237532606} {{TruthfulQA: Measuring How Models Mimic Human Falsehoods}}.
\newblock In \emph{Annual Meeting of the Association for Computational Linguistics}.

\bibitem[{Liu et~al.(2019)Liu, Ott, Goyal, Du, Joshi, Chen, Levy, Lewis, Zettlemoyer, and Stoyanov}]{liu2019roberta}
Yinhan Liu, Myle Ott, Naman Goyal, Jingfei Du, Mandar Joshi, Danqi Chen, Omer Levy, Mike Lewis, Luke Zettlemoyer, and Veselin Stoyanov. 2019.
\newblock {RoBERTa: A Robustly Optimized BERT Pretraining Approach}.
\newblock \emph{arXiv preprint arXiv:1907.11692}.

\bibitem[{Liu et~al.(2023)Liu, Deb, Teruel, Halfaker, Radev, and Awadallah}]{liu-etal-2023-improving}
Yixin Liu, Budhaditya Deb, Milagro Teruel, Aaron Halfaker, Dragomir Radev, and Ahmed~Hassan Awadallah. 2023.
\newblock \href {https://doi.org/10.18653/v1/2023.acl-long.844} {On improving summarization factual consistency from natural language feedback}.
\newblock In \emph{Proceedings of the 61st Annual Meeting of the Association for Computational Linguistics (Volume 1: Long Papers)}, pages 15144--15161, Toronto, Canada. Association for Computational Linguistics.

\bibitem[{Madaan et~al.(2023)Madaan, Tandon, Gupta, Hallinan, Gao, Wiegreffe, Alon, Dziri, Prabhumoye, Yang et~al.}]{madaan2024self}
Aman Madaan, Niket Tandon, Prakhar Gupta, Skyler Hallinan, Luyu Gao, Sarah Wiegreffe, Uri Alon, Nouha Dziri, Shrimai Prabhumoye, Yiming Yang, et~al. 2023.
\newblock {Self-refine: Iterative refinement with self-feedback}.
\newblock \emph{Advances in Neural Information Processing Systems}, 36.

\bibitem[{Meta(2024)}]{metallama3}
Meta. 2024.
\newblock \href {https://ai.meta.com/blog/meta-llama-3/} {{Introducing Meta Llama 3: The most capable openly available LLM to date}}.

\bibitem[{Mihaylov et~al.(2018)Mihaylov, Clark, Khot, and Sabharwal}]{Mihaylov2018CanAS}
Todor Mihaylov, Peter Clark, Tushar Khot, and Ashish Sabharwal. 2018.
\newblock \href {https://api.semanticscholar.org/CorpusID:52183757} {Can a suit of armor conduct electricity? a new dataset for open book question answering}.
\newblock In \emph{Conference on Empirical Methods in Natural Language Processing}.

\bibitem[{Mishra et~al.(2024)Mishra, Asai, Balachandran, Wang, Neubig, Tsvetkov, and Hajishirzi}]{mishra2024finegrained}
Abhika Mishra, Akari Asai, Vidhisha Balachandran, Yizhong Wang, Graham Neubig, Yulia Tsvetkov, and Hannaneh Hajishirzi. 2024.
\newblock \href {https://openreview.net/forum?id=dJMTn3QOWO} {Fine-grained hallucination detection and editing for language models}.
\newblock In \emph{First Conference on Language Modeling}.

\bibitem[{Nie et~al.(2020)Nie, Zhou, and Bansal}]{nie-etal-2020-learn}
Yixin Nie, Xiang Zhou, and Mohit Bansal. 2020.
\newblock \href {https://doi.org/10.18653/v1/2020.emnlp-main.734} {What can we learn from collective human opinions on natural language inference data?}
\newblock In \emph{Proceedings of the 2020 Conference on Empirical Methods in Natural Language Processing (EMNLP)}, pages 9131--9143, Online. Association for Computational Linguistics.

\bibitem[{Olausson et~al.(2024)Olausson, Inala, Wang, Gao, and Solar-Lezama}]{olausson2024is}
Theo~X. Olausson, Jeevana~Priya Inala, Chenglong Wang, Jianfeng Gao, and Armando Solar-Lezama. 2024.
\newblock \href {https://openreview.net/forum?id=y0GJXRungR} {Is self-repair a silver bullet for code generation?}
\newblock In \emph{The Twelfth International Conference on Learning Representations}.

\bibitem[{Pan et~al.(2024)Pan, Saxon, Xu, Nathani, Wang, and Wang}]{pan2024automatically}
Liangming Pan, Michael Saxon, Wenda Xu, Deepak Nathani, Xinyi Wang, and William~Yang Wang. 2024.
\newblock Automatically correcting large language models: Surveying the landscape of diverse automated correction strategies.
\newblock \emph{Transactions of the Association for Computational Linguistics}, 12:484--506.

\bibitem[{Panthaplackel et~al.(2021)Panthaplackel, Allamanis, and Brockschmidt}]{panthaplackel2021copy}
Sheena Panthaplackel, Miltiadis Allamanis, and Marc Brockschmidt. 2021.
\newblock Copy that! editing sequences by copying spans.
\newblock In \emph{Proceedings of the AAAI Conference on Artificial Intelligence}, volume~35, pages 13622--13630.

\bibitem[{Paul et~al.(2024)Paul, Ismayilzada, Peyrard, Borges, Bosselut, West, and Faltings}]{paul-etal-2024-refiner}
Debjit Paul, Mete Ismayilzada, Maxime Peyrard, Beatriz Borges, Antoine Bosselut, Robert West, and Boi Faltings. 2024.
\newblock \href {https://aclanthology.org/2024.eacl-long.67} {{REFINER}: Reasoning feedback on intermediate representations}.
\newblock In \emph{Proceedings of the 18th Conference of the European Chapter of the Association for Computational Linguistics (Volume 1: Long Papers)}, pages 1100--1126, St. Julian{'}s, Malta. Association for Computational Linguistics.

\bibitem[{Pavlick and Kwiatkowski(2019)}]{pavlick-kwiatkowski-2019-inherent}
Ellie Pavlick and Tom Kwiatkowski. 2019.
\newblock \href {https://doi.org/10.1162/tacl_a_00293} {Inherent disagreements in human textual inferences}.
\newblock \emph{Transactions of the Association for Computational Linguistics}, 7:677--694.

\bibitem[{Raunak et~al.(2023)Raunak, Sharaf, Wang, Awadalla, and Menezes}]{raunak-etal-2023-leveraging}
Vikas Raunak, Amr Sharaf, Yiren Wang, Hany Awadalla, and Arul Menezes. 2023.
\newblock \href {https://doi.org/10.18653/v1/2023.findings-emnlp.804} {Leveraging {GPT}-4 for automatic translation post-editing}.
\newblock In \emph{Findings of the Association for Computational Linguistics: EMNLP 2023}, pages 12009--12024, Singapore. Association for Computational Linguistics.

\bibitem[{Saunders(2023)}]{saunders-2023-improving}
Jarem Saunders. 2023.
\newblock \href {https://doi.org/10.18653/v1/2023.sigmorphon-1.10} {Improving automated prediction of {E}nglish lexical blends through the use of observable linguistic features}.
\newblock In \emph{Proceedings of the 20th SIGMORPHON workshop on Computational Research in Phonetics, Phonology, and Morphology}, pages 93--97, Toronto, Canada. Association for Computational Linguistics.

\bibitem[{Saunders et~al.(2022)Saunders, Yeh, Wu, Bills, Ouyang, Ward, and Leike}]{saunders2022self}
William Saunders, Catherine Yeh, Jeff Wu, Steven Bills, Long Ouyang, Jonathan Ward, and Jan Leike. 2022.
\newblock Self-critiquing models for assisting human evaluators.
\newblock \emph{arXiv preprint arXiv:2206.05802}.

\bibitem[{Shaikh et~al.(2023)Shaikh, Zhang, Held, Bernstein, and Yang}]{shaikh2023second}
Omar Shaikh, Hongxin Zhang, William Held, Michael Bernstein, and Diyi Yang. 2023.
\newblock On second thought, let’s not think step by step! bias and toxicity in zero-shot reasoning.
\newblock In \emph{Proceedings of the 61st Annual Meeting of the Association for Computational Linguistics (Volume 1: Long Papers)}, pages 4454--4470.

\bibitem[{Shinn et~al.(2024)Shinn, Cassano, Gopinath, Narasimhan, and Yao}]{shinn2024reflexion}
Noah Shinn, Federico Cassano, Ashwin Gopinath, Karthik Narasimhan, and Shunyu Yao. 2024.
\newblock Reflexion: Language agents with verbal reinforcement learning.
\newblock \emph{Advances in Neural Information Processing Systems}, 36.

\bibitem[{Shypula et~al.(2024)Shypula, Madaan, Zeng, Alon, Gardner, Yang, Hashemi, Neubig, Ranganathan, Bastani, and Yazdanbakhsh}]{shypula2024learning}
Alexander~G Shypula, Aman Madaan, Yimeng Zeng, Uri Alon, Jacob~R. Gardner, Yiming Yang, Milad Hashemi, Graham Neubig, Parthasarathy Ranganathan, Osbert Bastani, and Amir Yazdanbakhsh. 2024.
\newblock \href {https://openreview.net/forum?id=ix7rLVHXyY} {Learning performance-improving code edits}.
\newblock In \emph{The Twelfth International Conference on Learning Representations}.

\bibitem[{Stengel-Eskin et~al.(2024)Stengel-Eskin, Prasad, and Bansal}]{stengel2024regal}
Elias Stengel-Eskin, Archiki Prasad, and Mohit Bansal. 2024.
\newblock {ReGAL: Refactoring Programs to Discover Generalizable Abstractions}.
\newblock \emph{arXiv preprint arXiv:2401.16467}.

\bibitem[{Tang et~al.(2024{\natexlab{a}})Tang, Laban, and Durrett}]{tang2024minicheck}
Liyan Tang, Philippe Laban, and Greg Durrett. 2024{\natexlab{a}}.
\newblock {MiniCheck: Efficient Fact-Checking of LLMs on Grounding Documents}.
\newblock \emph{arXiv preprint arXiv:2404.10774}.

\bibitem[{Tang et~al.(2024{\natexlab{b}})Tang, Shalyminov, Wong, Burnsky, Vincent, Yang, Singh, Feng, Song, Su, Sun, Zhang, Mansour, and McKeown}]{Tang2024}
Liyan Tang, Igor Shalyminov, Amy Wong, Jon Burnsky, Jake Vincent, Yu’an Yang, Siffi Singh, Song Feng, Hwanjun Song, Hang Su, Justin Sun, Yi~Zhang, Saab Mansour, and Kathleen McKeown. 2024{\natexlab{b}}.
\newblock \href {https://www.amazon.science/publications/tofueval-evaluating-hallucinations-of-llms-on-topic-focused-dialogue-summarization} {{TofuEval: Evaluating hallucinations of LLMs on topic-focused dialogue summarization}}.
\newblock In \emph{NAACL 2024}.

\bibitem[{Thorne and Vlachos(2021)}]{thorne-vlachos-2021-evidence}
James Thorne and Andreas Vlachos. 2021.
\newblock \href {https://doi.org/10.18653/v1/2021.acl-long.256} {Evidence-based factual error correction}.
\newblock In \emph{Proceedings of the 59th Annual Meeting of the Association for Computational Linguistics and the 11th International Joint Conference on Natural Language Processing (Volume 1: Long Papers)}, pages 3298--3309, Online. Association for Computational Linguistics.

\bibitem[{Tian et~al.(2023)Tian, Xu, Zhang, and Mao}]{tian-etal-2023-dynamic}
Yuan Tian, Nan Xu, Ruike Zhang, and Wenji Mao. 2023.
\newblock \href {https://doi.org/10.18653/v1/2023.acl-long.139} {Dynamic routing transformer network for multimodal sarcasm detection}.
\newblock In \emph{Proceedings of the 61st Annual Meeting of the Association for Computational Linguistics (Volume 1: Long Papers)}, pages 2468--2480, Toronto, Canada. Association for Computational Linguistics.

\bibitem[{Touvron et~al.(2023)Touvron, Martin, Stone, Albert, Almahairi, Babaei, Bashlykov, Batra, Bhargava, Bhosale et~al.}]{touvron2023llama}
Hugo Touvron, Louis Martin, Kevin Stone, Peter Albert, Amjad Almahairi, Yasmine Babaei, Nikolay Bashlykov, Soumya Batra, Prajjwal Bhargava, Shruti Bhosale, et~al. 2023.
\newblock {Llama 2: Open foundation and fine-tuned chat models}.
\newblock \emph{arXiv preprint arXiv:2307.09288}.

\bibitem[{Trienes et~al.(2024)Trienes, Joseph, Schl{\"o}tterer, Seifert, Lo, Xu, Wallace, and Li}]{Trienes:2024:ACL}
Jan Trienes, Sebastian Joseph, J{\"o}rg Schl{\"o}tterer, Christin Seifert, Kyle Lo, Wei Xu, Byron~C. Wallace, and Junyi~Jessy Li. 2024.
\newblock {InfoLossQA}: {C}haracterizing and recovering information loss in text simplification.
\newblock In \emph{Proceedings of the 62nd Annual Meeting of the Association for Computational Linguistics}.

\bibitem[{Turpin et~al.(2024)Turpin, Michael, Perez, and Bowman}]{turpin2024language}
Miles Turpin, Julian Michael, Ethan Perez, and Samuel Bowman. 2024.
\newblock Language models don't always say what they think: unfaithful explanations in chain-of-thought prompting.
\newblock \emph{Advances in Neural Information Processing Systems}, 36.

\bibitem[{Wadhwa et~al.(2024)Wadhwa, Chen, Li, and Durrett}]{wadhwa2024using}
Manya Wadhwa, Jifan Chen, Junyi~Jessy Li, and Greg Durrett. 2024.
\newblock \href {https://openreview.net/forum?id=Qmq4zqdnWh} {Using natural language explanations to rescale human judgments}.
\newblock In \emph{First Conference on Language Modeling}.

\bibitem[{Wang et~al.(2024)Wang, Li, Chen, Cai, Zhu, Lin, Cao, Kong, Liu, Liu, and Sui}]{wang-etal-2024-large-language-models-fair}
Peiyi Wang, Lei Li, Liang Chen, Zefan Cai, Dawei Zhu, Binghuai Lin, Yunbo Cao, Lingpeng Kong, Qi~Liu, Tianyu Liu, and Zhifang Sui. 2024.
\newblock \href {https://doi.org/10.18653/v1/2024.acl-long.511} {Large language models are not fair evaluators}.
\newblock In \emph{Proceedings of the 62nd Annual Meeting of the Association for Computational Linguistics (Volume 1: Long Papers)}, pages 9440--9450, Bangkok, Thailand. Association for Computational Linguistics.

\bibitem[{Wang et~al.(2023)Wang, Yu, Tan, O'Brien, Pasunuru, Dwivedi-Yu, Golovneva, Zettlemoyer, Fazel-Zarandi, and Celikyilmaz}]{wang2023shepherd}
Tianlu Wang, Ping Yu, Xiaoqing~Ellen Tan, Sean O'Brien, Ramakanth Pasunuru, Jane Dwivedi-Yu, Olga Golovneva, Luke Zettlemoyer, Maryam Fazel-Zarandi, and Asli Celikyilmaz. 2023.
\newblock {Shepherd: A critic for language model generation}.
\newblock \emph{arXiv preprint arXiv:2308.04592}.

\bibitem[{Welleck et~al.(2023)Welleck, Lu, West, Brahman, Shen, Khashabi, and Choi}]{welleck2023generating}
Sean Welleck, Ximing Lu, Peter West, Faeze Brahman, Tianxiao Shen, Daniel Khashabi, and Yejin Choi. 2023.
\newblock \href {https://openreview.net/forum?id=hH36JeQZDaO} {Generating sequences by learning to self-correct}.
\newblock In \emph{The Eleventh International Conference on Learning Representations}.

\bibitem[{Xu et~al.(2024)Xu, Deutsch, Finkelstein, Juraska, Zhang, Liu, Wang, Li, and Freitag}]{xu-etal-2024-llmrefine}
Wenda Xu, Daniel Deutsch, Mara Finkelstein, Juraj Juraska, Biao Zhang, Zhongtao Liu, William~Yang Wang, Lei Li, and Markus Freitag. 2024.
\newblock \href {https://doi.org/10.18653/v1/2024.findings-naacl.92} {{LLMR}efine: Pinpointing and refining large language models via fine-grained actionable feedback}.
\newblock In \emph{Findings of the Association for Computational Linguistics: NAACL 2024}, pages 1429--1445, Mexico City, Mexico. Association for Computational Linguistics.

\bibitem[{Xu et~al.(2023)Xu, Wang, Pan, Song, Freitag, Wang, and Li}]{xu-etal-2023-instructscore}
Wenda Xu, Danqing Wang, Liangming Pan, Zhenqiao Song, Markus Freitag, William Wang, and Lei Li. 2023.
\newblock \href {https://doi.org/10.18653/v1/2023.emnlp-main.365} {{INSTRUCTSCORE}: Towards explainable text generation evaluation with automatic feedback}.
\newblock In \emph{Proceedings of the 2023 Conference on Empirical Methods in Natural Language Processing}, pages 5967--5994, Singapore. Association for Computational Linguistics.

\bibitem[{Ye et~al.(2023)Ye, Jo, Kim, Kim, Hwang, and Seo}]{selfee2023}
Seonghyeon Ye, Yongrae Jo, Doyoung Kim, Sungdong Kim, Hyeonbin Hwang, and Minjoon Seo. 2023.
\newblock \href {https://kaistai.github.io/SelFee/} {{SelFee: Iterative Self-Revising LLM Empowered by Self-Feedback Generation}}.
\newblock Blog post.

\bibitem[{Ye and Durrett(2022)}]{Ye-Durrett:2022:Fewshot}
Xi~Ye and Greg Durrett. 2022.
\newblock {The Unreliability of Explanations in Few-shot Prompting for Textual Reasoning}.
\newblock In \emph{Advances in Neural Information Processing Systems}.

\bibitem[{Zha et~al.(2023)Zha, Yang, Li, and Hu}]{zha-etal-2023-alignscore}
Yuheng Zha, Yichi Yang, Ruichen Li, and Zhiting Hu. 2023.
\newblock \href {https://doi.org/10.18653/v1/2023.acl-long.634} {{A}lign{S}core: Evaluating factual consistency with a unified alignment function}.
\newblock In \emph{Proceedings of the 61st Annual Meeting of the Association for Computational Linguistics (Volume 1: Long Papers)}, pages 11328--11348, Toronto, Canada. Association for Computational Linguistics.

\bibitem[{Zhang et~al.(2024)Zhang, Press, Merrill, Liu, and Smith}]{zhang2024how}
Muru Zhang, Ofir Press, William Merrill, Alisa Liu, and Noah~A. Smith. 2024.
\newblock \href {https://openreview.net/forum?id=FPlaQyAGHu} {How language model hallucinations can snowball}.
\newblock In \emph{Forty-first International Conference on Machine Learning}.

\bibitem[{Zhu et~al.(2021)Zhu, Liu, Mei, and Zeng}]{zhu-etal-2021-mediasum}
Chenguang Zhu, Yang Liu, Jie Mei, and Michael Zeng. 2021.
\newblock \href {https://doi.org/10.18653/v1/2021.naacl-main.474} {{M}edia{S}um: A large-scale media interview dataset for dialogue summarization}.
\newblock In \emph{Proceedings of the 2021 Conference of the North American Chapter of the Association for Computational Linguistics: Human Language Technologies}, pages 5927--5934, Online. Association for Computational Linguistics.

\bibitem[{Zhuo et~al.(2023)Zhuo, Huang, Chen, and Xing}]{zhuo2023red}
Terry~Yue Zhuo, Yujin Huang, Chunyang Chen, and Zhenchang Xing. 2023.
\newblock {Red teaming ChatGPT via Jailbreaking: Bias, Robustness, Reliability and Toxicity}.
\newblock \emph{arXiv preprint arXiv:2301.12867}.

\end{thebibliography}

\appendix

\section{Dataset Creation} \label{appen:data_creation}

\subsection{UltraChat}

UltraChat is an open-source, large-scale, and multi-round dialogue data, released under the MIT License. The dataset was constructed with the aim of constructing powerful language models with general conversational capability.

As mentioned in Section \ref{sec:datasets}, we use a subset of document-grounded instructions from UltraChat. To guarantee sufficient context, we only sample instances which are at least 1000 characters long. Each UltraChat instance has a source document and the summarization instruction in one prompt. To insure that the instruction is summarization related, we check for the following list of keywords: [``can you summarize'',``summarize the following'',``give a summary'',``can you provide a summary'',``provide a brief summary'',``summarize the'',``can you give me a summary'']. This dataset is in English only. Tables \ref{tab:ultrachat_examples} and   \ref{tab:ultrachat_example_2} show examples of UltraChat instructions.

\subsection{MediaSum/TofuEval}

TofuEval is a benchmark evaluating factual consistency of document grounded summaries. This dataset is released under the MIT-0 license. 

We use the MediaSum subset of TofuEval as our evaluation set; but sample from MediaSum's original train set to gather more data for training. We create a summary instruction process similar to TofuEval, where they create topic-focused summaries by prompting GPT-4 to generate 3 topics being discussed in the dialogue to be summarized, and then converting each of those topics to be a summarization instruction. We follow the same generation strategy and prompt GPT-4 (Prompt \ref{prompt:topic_generation}) to give 3 topics under discussion in the sampled dialogues from MediaSum. We then use topics as separate summarization instruction and generate an initial response. Tables \ref{tab:mediasum_example_1} and   \ref{tab:mediasum_example_2} show examples of MediaSum source document along with the instruction.

This dataset is in English only.

\begin{prompt}[title={\footnotesize\texttt{Prompt \thetcbcounter: Zero-Shot prompt used with GPT-4 to generate topics for MediaSum articles}}, label=prompt:topic_generation]
Document: \\
\{\{ \paramnorm{document} \}\}\\
Enumerate three main topics that people would like to know from the provided document. Each topic should be around 5 words.
\end{prompt}

\subsection{Initial Response Generation}\label{subsec:initial_response}

We prompt \textsc{Llama2-7b-Chat} to get responses for instructions from UltraChat and MediaSum. Since UltraChat instructions have the source document and the summarization instruction, we use them as is. For MediaSum, we create a summarization prompt using the topics generated by GPT-4. The initial response generation prompt for MediaSum is given in prompt \ref{prompt:initial_response_mediasum}. Tables \ref{tab:ultrachat_examples} and \ref{tab:ultrachat_example_2} show examples of the UltraChat prompt along with the initial response. Tables \ref{tab:mediasum_example_1} and \ref{tab:mediasum_example_2} show examples of the MediaSum prompt along with initial response.

Once we get initial responses for UltraChat, we filter and only keep instances where the initial response length is shorter than the source document.

\begin{prompt}[title={\footnotesize\texttt{Prompt \thetcbcounter: Prompt used with \textsc{Llama2-7b-chat} to generate the initial summary for MediaSum instances}}, label=prompt:initial_response_mediasum]
Document: \\
\{\{ \paramnorm{document} \}\}\\
Summarize the provided document focusing on \{\{\paramnorm{topic}\}\}. The summary should be less than 50 words in length.
\end{prompt}

\section{Supervised Fine-Tuning}\label{append:ft}

In Section \ref{sec:fine_tuning} we outline the creation of the training data and its use for fine-tuning. The structured feedback from GPT-4 has the following components: error localization in the form of a span, feedback reasoning for why the span is inconsistent and a suggested fix. We map this structured feedback to a natural language form using the following template: \textcolor{purple}{\{feedback\} The error span is: `\{span\}'. To fix this, consider changing the span to `\{fix\}'"}

 We fine-tune  \textsc{Llama-3-8b-Instruct} and \textsc{Llama-2-7b-chat} for our proposed method as well as for the baselines. We use SFTTrainer from TRL \cite{trl} to fine-tune. All the fine-tuning can be conducted on 3 x A6000 GPU with 48GB of memory. We use LoRA \cite{hu2022lora} (Rank = 8) with a learning rate of  2e-4 and a warmup ratio of 0.05. We set the per GPU batch size to 2 and the maximum sequence length to 2048.

\begin{prompt}[title={\footnotesize\texttt{Prompt \thetcbcounter: $p_\text{critique}$ used to generate feedback for training data creation from GPT-4 }}, label=prompt:critique_teacher]
Summarize the following document on the topic: \{\{\paramnorm{topic}\}\}:
\{\{\paramnorm{document}\}\}

Summary on topic: \{\{\paramnorm{topic}\}\}

\{\{\paramnorm{summary}\}\}

-----

The provided summary is factually inconsistent with the corresponding document. This implies that there is information in the summary that is NOT substantiated by the document. Factual inconsistencies can be of the following types:\\
1. Mis-Referencing: a property or an event in the summary can be found in the document, but are associated with the wrong entity \\
2. Stating Opinion As Fact: the summary entails a proposition that's mentioned in the document not as a fact, but as someone's opinion \\  
3. Reasoning Error: the summary makes one or more wrong inferences from the information in the document\\
4. Tense/modality Error: the tense or modal (eg: can, may, must) used in the summary does not match the tense/modality of the document \\ 
5. Extrinsic Information: the summary contains new information not grounded in the source document \\ 
6. Contradiction: the summary contradicts the document \\ 
7. Nuanced Meaning Shift: the summary twists information from the document in a subtle way \\

Identify factually inconsistent information in the form of a JSON and return a list with the following keys: \\ 
1. inconsistency: <span from the summary that is factually inconsistent>\\
2. inconsistency type: <the inconsistency type from the above list of types>\\
3. feedback: <explanation of the error and how it can be fixed>\\
4. fix: <correct span that fixes the inconsistency>\\
\end{prompt}

\begin{prompt}[title={\footnotesize\texttt{Prompt \thetcbcounter:  $p_\text{refine}$ to generate refinements with natural language feedback for training data curation}}, label=prompt:refine_teacher]
I summarized the following document on the topic: '\{\{\paramnorm{topic}\}\}:
\{\{\paramnorm{document}\}\} \\

Summary of the above document on topic: \{\{\paramnorm{topic}\}\}:
\{\{\paramnorm{summary}\}\} \\ 

Feedback for the above summary: 
\{\{\paramnorm{feedback}\}\} \\

Edit the summary such that the refinement doesn't have any errors mentioned in the feedback. Make the minimum number of changes when doing the refinement.
\end{prompt}

\section{Refinement Strategies}\label{append:refinement_prompts}
We list all instructions for the different refinement strategies described in Section \ref{subsec:baselines}.

\paragraph{One-Step: Direct Refinement (DR)} We prompt the model with a general refinement instruction to improve the factual consistency of the initial response. This strategy does not use any external or self-generated feedback. 

Prompt Instruction:

\begin{prompt}[title={\footnotesize\texttt{Prompt \thetcbcounter: DR Prompt for MediaSum}}, label=prompt:dr]
I summarized the following document on the topic: `\{\{\paramnorm{topic}\}\}':\\
\{\{\paramnorm{document}\}\} \\

Summary of the above document on topic `\{\{\paramnorm{topic}\}\}':
\{\{\paramnorm{summary}\}\} \\ 
\\
If there are any factual inconsistencies in the summary then edit the summary such that the refinement doesn't have any inconsistencies. Consistency in this context implies that all information presented in the summary is substantiated by the document.If the summary is consistent, then just the copy the same summary with no changes. When refining, make the minimum number of changes.
\end{prompt}

\paragraph{Two-Step: Refinement with Natural Language Feedback (Feed+DR)} We first prompt the model to generate a feedback reasoning about any factual inconsistencies in the initial response, then we pass the feedback to the refinement model. The feedback in this case is generated sentence wise, combined together and then used for refinement.

Feedback Prompt Instruction: 

\begin{prompt}[title={\footnotesize\texttt{Prompt \thetcbcounter: Feedback Prompt for MediaSum For Two-Step refinement}}, label=prompt:feedback_two_step]
I summarized the following document on the topic: `\{\{\paramnorm{topic}\}\}':\\
\{\{\paramnorm{document}\}\} \\

Summary of the above document on topic `\{\{\paramnorm{topic}\}\}':
\{\{\paramnorm{summary}\}\} \\ 
\\

For the following sentence in the summary:\\
\{\{\paramnorm{sentence}\}\} \\ 

reason if there is any factually inconsistent span in the sentence. A span is factually inconsistent if it cannot be substantiated by the document. If there is no inconsistency, then end your answer with ``no error''. Otherwise if there is a factual inconsistency, then give reasons for it, point to the error span by stating ``The error span: <span from sentence>'' and end your answer with a suggested fix to the summary
\end{prompt}

Refinement Prompt Instruction:
\begin{prompt}[title={\footnotesize\texttt{Prompt \thetcbcounter: Refine Prompt for MediaSum For Two-Step refinement}}, label=prompt:refine_two_step]
I summarized the following document on the topic: `\{\{\paramnorm{topic}\}\}':\\
\{\{\paramnorm{document}\}\} \\

Summary of the above document on topic `\{\{\paramnorm{topic}\}\}':
\{\{\paramnorm{summary}\}\} \\ 

Feedback for the above summary: 
\{\{\paramnorm{feedback}\}\} \\ 
\\
Edit the user response such that the refinement doesn't have any errors mentioned in the feedback. Make the minimum number of changes when doing the refinement. Do not include a preamble.
\end{prompt}

\paragraph{Two-Step: Direct Refinement with \textsc{Detect} (Detect+DR)} We first filter any initial responses that are factually consistent using our $M_\text{detect}$, and then refine the summaries that have an inconsistency with a general instruction. This baseline does not use any external or self-generated feedback.

Prompt Instruction:

\begin{prompt}[title={\footnotesize\texttt{Prompt \thetcbcounter: Refinement prompt for MediaSum for Two-Step: Direct Refinement with \textsc{Detect} }}, label=prompt:refine_two_step_with_dr]
I summarized the following document on the topic: `\{\{\paramnorm{topic}\}\}':\\
\{\{\paramnorm{document}\}\} \\

Summary of the above document on topic `\{\{\paramnorm{topic}\}\}':
\{\{\paramnorm{summary}\}\} \\ 
\\
Edit the response such that the refinement doesn't have any factual inconsistencies. Consistency in this context implies that all information presented in the response is substantiated by the document. When refining, make the minimum number of changes.
\end{prompt}

Note: in this baseline we remove the need for the model to do the detection and refinement together. 

\paragraph{Three-Step: \textsc{Detect}, \textsc{Reason}, \textsc{Fix}: DCR (proposed)} We first filter any initial responses that are factually consistent using the $M_\text{detect}$. We use $M_\text{critique}$ to generate a sentence wise feedback for any sentences that were detected to have an error. We combine this feedback and generate a refinement using $M_\text{refine}$.

Feedback Prompt:
\begin{prompt}[title={\footnotesize\texttt{Prompt \thetcbcounter: Feedback prompt for MediaSum for Three-Step Refinement:DCR (proposed) }}, label=prompt:feedback_three_step]
I summarized the following document on the topic: `\{\{\paramnorm{topic}\}\}':\\
\{\{\paramnorm{document}\}\} \\

Summary of the above document on topic `\{\{\paramnorm{topic}\}\}':
\{\{\paramnorm{summary}\}\} \\ 
\\
reason about the factually inconsistent span in the sentence. A span is factually inconsistent if it cannot be substantiated by the document. Give reasons for the factual inconsistency, point to the error span by stating ``The error span: <span from sentence> and end your answer with a suggested fix to the summary.
\end{prompt}

Refinement Instruction:

\begin{prompt}[title={\footnotesize\texttt{Prompt \thetcbcounter: Refinement prompt for MediaSum for Three-Step Refinement:DCR (proposed) }}, label=prompt:refine_three_step]
I summarized the following document on the topic: `\{\{\paramnorm{topic}\}\}':\\
\{\{\paramnorm{document}\}\} \\

Summary of the above document on topic `\{\{\paramnorm{topic}\}\}':
\{\{\paramnorm{summary}\}\} \\ 
Feedback for the above summary: 
\{\{\paramnorm{feedback}\}\} \\ 
\\
Edit the user response such that the refinement doesn't have any errors mentioned in the feedback. Make the minimum number of changes when doing the refinement. Do not include a preamble.
\end{prompt}

\section{Automatic Evaluation}
\subsection{GPT-4 Factuality Likert Scale }\label{appen:eval_gpt4}

\begin{prompt}[title={\footnotesize\texttt{Prompt \thetcbcounter: Zero-Shot prompt used with GPT-4 to generate a factuality score on a scale of 1-5.}}, label=prompt:gpt4_eval_likert]
Determine whether the provided summary is consistent with the corresponding document. Consistency in this context implies that all information presented in the response is substantiated by the document. If not, it should be considered inconsistent.\\
\{\{ \paramnorm{instruction} \}\}\\
\{\{ \paramnorm{response} \}\}\\
The response can have one or more of the following errors:\\
1. Extrinsic Information: the response contains new information not grounded in the source material\\
2. Mis-Referencing: a property or an event in the response can be found in the source material, but are associated with the wrong entity \\
3. Stating Opinion As Fact: the response entails a proposition that's mentioned in the source material not as a fact, but as someone's opinion \\
4. Reasoning Error: the response makes one or more wrong inferences from the information in the source material \\
5. Tense/modality Error: the tense or modal (eg: can, may, must) used in the response sentence does not match the tense/modality of the source material \\
6. Contradiction: the response contradicts the source material \\
7. Nuanced Meaning Shift: the response twists information from the source material in a subtle way \\
Given the error categories, rate the above response on a scale of 1 to 5 based on extent of factual consistency:\\
\\
5. completely consistent: the response is completely factually consistent with the source material.\\
4. insignificant inconsistencies:  the response is mostly factually consistent, with slight inconsistencies not affecting main points. \\
3. partially inconsistent: overall factually consistent, with a few inconsistencies with the source material.\\
2. severe inconsistencies: nearly half response is factually inconsistent, with severe deviation from main points.\\
1. completely inconsistent: the entire response is factually inconsistent with the source material.\\
First output a list of errors that the summary makes, then conclude the response with a score in the following format: "therefore, the score is:"
\end{prompt}

\subsection{GPT-4-Pairwise Evaluation For Win-Rate}\label{appen:eval_gpt4_pairwise}

\begin{prompt}[title={\footnotesize\texttt{Prompt \thetcbcounter: Zero-Shot prompt used with GPT-4 to generate pairwise factuality scores on a scale of 1-5.}}, label=prompt:gpt4_pairwise_eval]
Determine whether the provided summary is consistent with the corresponding document. Consistency in this context implies that all information presented in the response is substantiated by the document. If not, it should be considered inconsistent. \\
\{\{ \paramnorm{instruction} \}\}\\
\#\# Response 1
\{\{ \paramnorm{response1} \}\}\\

\#\# Response 2
\{\{ \paramnorm{response2} \}\}\\

A response can have one or more of the following errors:\\
1. Extrinsic Information: the response contains new information not grounded in the source material\\
2. Mis-Referencing: a property or an event in the response can be found in the source material, but are associated with the wrong entity\\
3. Stating Opinion As Fact: the response entails a proposition that's mentioned in the source material not as a fact, but as someone's opinion\\
4. Reasoning Error: the response makes one or more wrong inferences from the information in the source material \\
5. Tense/modality Error: the tense or modal (eg: can, may, must) used in the response sentence does not match the tense/modality of the source material \\ 
6. Contradiction: the response contradicts the source material \\
7. Nuanced Meaning Shift: the response twists information from the source material in a subtle way\\

Given the error categories, rate each response on a scale of 1 to 5 based on extent of factual consistency: \\ 
5. completely consistent: the response is completely factually consistent with the source material. \\
4. insignificant inconsistencies:  the response is mostly factually consistent, with slight inconsistencies not affecting main points.\\
3. partially inconsistent: overall factually consistent, with a few inconsistencies with the source material.\\
2. severe inconsistencies: nearly half response is factually inconsistent, with severe deviation from main points.\\
1. completely inconsistent: the entire response is factually inconsistent with the source material\\
For each response, first output a list of errors that the summary makes, then conclude the response with a score in the following format: "therefore, the score is:"\\
\\
Output Format:\\
\\
\#\# Response 1\\
...\\
\\
\#\# Response 2 \\ 
...\\
\\
\end{prompt}

\section{\textsc{Detect} Step}
We use MiniCheck \cite{tang2024minicheck} as $M_\text{detect}$. To check its performance against ground truth binary factual consistency labels, we calculate precision/recall/F1 and balanced accuracy on the MediaSum split of TofuEval which has sentence level factual consistency labels. MiniCheck achieves a sentence-level balanced accuracy of 73.6\%. It achieves a precision of 0.54 and recall of 0.64 on the task of detecting factually inconsistent sentences.

\section{\textsc{Critique} Step}

\begin{prompt}[title={\footnotesize\texttt{Prompt \thetcbcounter: Zero-Shot prompt used with GPT-4 to compare ground truth human written feedback against model generated feedback}}, label=prompt:gt_comparison]
Document: \\

Sentence: \{\{ \paramnorm{sentence} \}\}\\

For the above sentence, I received the following two feedbacks:\\
Feedback 1:\\
\{\{ \paramnorm{feedback1} \}\}\\
\\
Feedback 2:\\
\{\{ \paramnorm{feedback2} \}\}\\
\\
Are feedback 1 and feedback 2 talking about the same error in the sentence? Respond with one of the following: \\
(1) same error or mostly the same error, one of them covers a broader range of errors 
(2) totally different errors 
(3) feedback 2 says there is no error but feedback 1 has an error mentioned
\end{prompt}

\section{\textsc{Refine} Step} \label{append:refine}
As mentioned in Section \ref{subsec:baselines}, we train our baselines to be comparable to our proposed method. However, we note that fine-tuning DR baselines leads to the model learning optimize for the ``easy'' action and learning to copy instead of making fine-grained edits.  We observe a similar behavior when training $M_\text{critique}$ with a balanced set of sentences for Feed+Refine. The model learns the easier generation and learns to predict ``no error''. Prior work \cite{tian-etal-2023-dynamic,panthaplackel2021copy} has observed this behavior with seq2seq models, and we leave further exploration of this to future work.

Due to this behavior of our trained baselines, we observe that the fine-tuned baselines edit fewer summaries. Table \ref{tab:percentage_summ_changed} shows the percentage of summaries that remain unchanged in our test set. When refining with DR, with Refine-L3-FT as $M_\text{refine}$, the model only edits ~20\% of the responses and Refine-L2-FT only edits 5\% of responses. Similarly, when refining with Feed+DR, (Critique-L3-FT, Refine-L3-FT)  edits 26\% of the responses while (Critique-L2-FT, Refine-L2-FT) edits 19\% of responses. 

Using Levenshtein distance \cite{levenshtein1966binary}, we calculate the number of edits between the refinement and the initial response.  We break down the edit distance by the number of deletes, adds and substitutions. Table \ref{tab:edits} shows the average number of edits made by different models when refining with DCR. Note, the number of edits are averaged over \textbf{only} edited summaries. For each model, we see that the largest edit is the ``substitution (sub)'' operation. While the word level edits are some times ~50\% of the original length of the response, in Table \ref{tab:feedback_refinement_examples} we can qualitatively see that the refinements preserve the semantics, style and structure of the initial response and follow the feedback.

\section{Data Release}
We will release all our data, code and models under the MIT License.

\begin{table}[t!]
\centering
\small
\renewcommand{\tabcolsep}{0.8mm}
\begin{tabular}{l|l|rrr}
\toprule
\makecell{Error\\Granularity} & Edit Type & \makecell{GPT-4\\ \%} & \makecell{Refine-\\L3-FT \%} & \makecell{Refine-\\L2-FT \%} \\
\midrule
Word & Delete   & 4.0  & 4.0   & 0.0   \\
& Insert  & 2.0   & 2.0    & 4.0  \\
& Substitute    & 0.0    & 6.0   & 2.0   \\
\midrule
Phrase & Delete  & 4.0 & 6.0  & 10.0   \\
& Insert   & 2.0 & 12.0   & 8.0   \\
& Substitute   & 24.0   & 40 .0  & 28.0  \\
\midrule
Sentence & Delete & 2.0 & 2.0    & 4.0  \\
& Substitute & 16.0     & 16.0   & 38.0 \\
\midrule
  Phrase/Phrase & \makecell[l]{Substitute/ \\Insert}   & 10.0   & 6.0  & 0.0 \\
& \makecell[l]{Substitute/ \\Delete}   &0.0     &0.0   & 2.0\\
\midrule
\makecell[l]{Phrase/\\Sentence} & \makecell[l]{Substitute + \\Insert} & 14.0 & 4.0 & 0.0      \\
& \makecell[l]{Delete/ \\Insert}     & 2.0 & 0.0 & 0.0       \\
\midrule
No change & & 20.0 & 2.0 & 4.0 \\
\bottomrule
\end{tabular}
\vspace{-0.5em}
\caption{Distribution of different edit actions for model generated feedback. We a variation in the  granularity at which the errors are detects as well as the type of edits that the model feedback suggests.}
\label{tab:edit_distribution}
\vspace{-0.5em}
\end{table}

\begin{table*}[t!]
\centering 
\small
\rowcolors{3}{white}{lightgray} 
\renewcommand{\tabcolsep}{0.8mm}
\begin{tabular}{ll|ccc}
\toprule
& & \multicolumn{3}{c}{$M_\text{critique}$ / $M_\text{refine}$} \\
\midrule
Dataset & \makecell{Refinement Strategy} & \makecell{GPT-4/ GPT-4} & \makecell{Critique-L3-FT/ Refine-L3-FT} & \makecell{Critique-L2-FT/ Refine-L2-FT} \\
\midrule
TofuEval    & DR   & 0.66   & 0.83    & 0.95    \\
  & Feed+DR   & 0.08  & 0.74  & 0.81 \\
 & Detect+DR  & 0.49  & 0.75   & 0.65  \\
    & DCR   & 0.51  & 0.50 & 0.50  \\
   \midrule
UltraChat  & DR  & 0.87   & 0.91  & 0.99 \\
  & Feed+DR   & 0.04  & 0.85  & 0.91  \\
    & Detect+DR   & 0.24       & 0.84  & 0.86   \\
    & DCR   & 0.25  & 0.25  & 0.24 \\          \bottomrule
\end{tabular}
\caption{Percentage of responses that remain unchanged during the refinement process by different refinement baselines and DCR.}
\label{tab:percentage_summ_changed}
\end{table*}

\begin{table*}[t!]
\centering
\small
\renewcommand{\tabcolsep}{1.7mm}
\rowcolors{3}{white}{lightgray} 
\begin{tabular}{l|lll|cccccc}
\toprule
\multirow{2}{*}{Strategy} & \multirow{2}{*}{\textbf{$M_\text{detect}$}} & 
  \multirow{2}{*}{\textbf{$M_\text{critique}$}} &
  \multirow{2}{*}{\textbf{$M_\text{refine}$}} & 
  \multicolumn{6}{c}{MediaSum/TofuEval} \\
 & & & & 
\multicolumn{1}{c}{$A(r)$} & 
\multicolumn{1}{c}{$A(\hat{r})$} 
& $G(r)$ 
& $G(\hat{r})$ 
&  \makecell{$G(r)$\\pairwise} &  \makecell{$G(\hat{r})$\\pairwise} \\ 
\midrule
DR & - & - & Refine-L3-FT & 0.76&	0.80	&4.47&	4.67&4.45	&4.65  \\ 
Feed+DR & - & Critique-L3-FT & Refine-L3-FT &   0.76&	0.77	&4.45	&4.49	&4.49&	4.53\\ 
\textsc{Detect} + DR & MiniCheck & - & Refine-L3-FT & 0.76&	0.81	&4.47	&4.66&	4.45	&4.64  \\ 
DCR & MiniCheck & Critique-L3-FT & Refine-L3-FT & 0.76&	0.83	&4.45&	4.79&	4.41&	4.73 \\ 
\midrule
DR & - & - & Refine-L2-FT & 0.76&	0.77&	4.46	&4.46&	4.48&	4.50  \\ 
Feed+DR & - & Critique-L2-FT & Refine-L2-FT & 0.76	&0.75&	4.47	&4.43&	4.51	&4.52  \\ 
\textsc{Detect} + DR & MiniCheck & - & Refine-L2-FT & 0.76	&0.8&	4.48&	4.70&	4.49&	4.62  \\ 
DCR & MiniCheck & Critique-L2-FT & Refine-L2-FT & 0.76	&0.85&	4.49&	4.64	&4.43&	4.63 \\ 
\midrule 
DR & - & - & GPT-4 & 0.76	&0.78&	4.47&	4.69	&4.44&	4.69  \\ 
Feed+DR & - & GPT-4 & GPT-4 & 0.76	&0.86	&4.46&	4.95	&4.33	&4.82  \\ 
\textsc{Detect} + DR & MiniCheck & - & GPT-4 & 0.76	&0.87&	4.48&	4.94	&4.41&	4.88  \\ 
DCR & MiniCheck & GPT-4 & GPT-4 & 0.76&	0.86&	4.44&	4.97&	4.43	&4.88  \\ 
\bottomrule
\end{tabular}
\caption{Absolute metric values for different refinement strategies for TofuEval. $r$ is the original response, $\hat{r}$ is the refined response. $A$ is the AlignScore between 0-1. $G$ is the GPT4 likert score on 1-5. $G$ pairwise is the pairwise score of the original and refined response on a scale of 1-5.}
\label{tab:absolute_results_mediasum_1}
\end{table*}

\begin{table*}[t!]
\centering
\small
\renewcommand{\tabcolsep}{1.5mm}
\rowcolors{3}{white}{lightgray} 
\begin{tabular}{l|lll|cccccc}
\toprule
\multirow{2}{*}{Strategy} & \multirow{2}{*}{\textbf{$M_\text{detect}$}} & 
  \multirow{2}{*}{\textbf{$M_\text{critique}$}} &
  \multirow{2}{*}{\textbf{$M_\text{refine}$}} & 
  \multicolumn{6}{c}{UltraChat} \\
 & & & & 
\multicolumn{1}{c}{$A(r)$} & 
\multicolumn{1}{c}{$A(\hat{r})$} 
& $G(r)$ 
& $G(\hat{r})$ 
&  \makecell{$G(r)$\\pairwise} &  \makecell{$G(\hat{r})$\\pairwise} \\ 
\midrule
DR & - & - & Refine-L3-FT & 0.70&	0.71	&4.51&	4.59&	4.33	&4.43 \\ 
Feed+DR & - & Critique-L3-FT & Refine-L3-FT &   0.70&	0.70&	4.50	&4.52	&4.33	&4.33 \\ 
\textsc{Detect} + DR & MiniCheck & - & Refine-L3-FT & 0.70&	0.70	&4.46&	4.53&	4.27&	4.31 \\ 
DCR & MiniCheck & Critique-L3-FT & Refine-L3-FT & 0.70 &	0.75 &	4.50 &	4.54	& 4.23	& 4.32 \\ 
\midrule
DR & - & - & Refine-L2-FT & 0.69&	0.69	&4.48&	4.51&	4.30&	4.30  \\ 
Feed+DR & - & Critique-L2-FT & Refine-L2-FT & 0.7&	0.7	&4.51&	4.53&	4.32&	4.31  \\ 
\textsc{Detect} + DR & MiniCheck & - & Refine-L2-FT & 0.7	&0.71	&4.54&	4.56&	4.35&	4.38  \\ 
DCR & MiniCheck & Critique-L2-FT & Refine-L2-FT & 0.7&	0.73&	4.51	&4.42	&4.27&	4.21 \\ 
\midrule 
DR & - & - & GPT-4 & 0.70 &	0.71	&4.51	&4.59	&4.30&	4.39 \\ 
Feed+DR & - & GPT-4 & GPT-4 & 0.70&	0.79	&4.51&	4.79	&4.21	&4.47  \\ 
\textsc{Detect} + DR & MiniCheck & - & GPT-4 & 0.70	&0.74	&4.54&	4.81&	4.31&	4.60  \\ 
DCR & MiniCheck & GPT-4 & GPT-4 & 0.70	&0.76	&4.53	&4.74	&4.33	&4.58 \\ 
\bottomrule
\end{tabular}
\caption{Absolute metric values for different refinement strategies for UltraChat. $r$ is the original response, $\hat{r}$ is the refined response. $A$ is the AlignScore between 0-1. $G$ is the GPT4 likert score on 1-5. $G$ pairwise is the pairwise score of the original and refined response on a scale of 1-5.}
\label{tab:absolute_results_ultrachat_1}
\end{table*}

\begin{table*}[t!]
\centering
\small
\renewcommand{\tabcolsep}{1mm}
\begin{tabular}{l|ccccc|ccccc}
\toprule
& \multicolumn{5}{c|}{MediaSum/TofuEval} & \multicolumn{5}{c}{UltraChat}\\ 
\midrule
\makecell{$M_\text{critique}$/ $M_\text{refine}$} & Adds & Deletes & Subs & Len($r$) & Len($\hat{r}$) & Adds & Deletes & Subs & Len($r$) & Len($\hat{r}$) \\
\midrule
\makecell{Critique-L3-FT/ Refine-L3-FT}  &  5.0  & 7.4 & 9.3   & 53.3   & 50.9  & 6.6  & 42.4   & 13.3  & 233.5    & 197.8  \\
\makecell{Critique-L2-FT/ Refine-L2-FT} &  18.6   & 8.4  & 17.2  & 53.2   & 63.3    & 25.6  & 42.4   & 65.0   & 234.2 & 216.9   \\  
\makecell{GPT-4/GPT-4} &  11.8  & 5.0 & 15.9  & 53.3 & 60.1  & 11.6  & 44.1 & 55.4  & 234.7  & 202.2   \\
\bottomrule
\end{tabular}
\caption{Average edit distance broken down by the average number of additions, deletions, and substitutions between the refinement ($\hat{r}$) and original response ($r$) when refining with DCR. The values are averaged over edited summaries only. Table \ref{tab:percentage_summ_changed} shows the \% of summaries that remain unchanged by different refinement strategies including DCR.}
\label{tab:edits}
\end{table*}

\begin{table*}[t!]
\centering
\small
\begin{tabular}{p{15cm}}
\toprule
\multicolumn{0}{c}{\textbf{UltraChat Instruction - Example 1}} \\
\midrule
\textbf{Summarize the issues faced by Native American tribes on reservations in the mid-1800s, including lack of medical care and access to food, and how Indian agents attempted to address these problems.
Generate according to:} When the tribes got to the reservation in 1856, the federal Indian agents were then 100\% responsible for feeding them and caring for their health needs. I have documented in numerous essays that the federal government was slow to appropriate funds for the reservation, even when they had treaties, and that hunger and starvation was a major issue on the reservations. As well, I have documented that illnesses and diseases were also a major problem on the reservations. For the over 2000 Indians at Grand Ronde, there was only one doctor, and while there was a hospital it was limited, because in the first few years some 75\% of the Indians were sick in any 6 month period. Many people died on the reservation without any medical attention, either because they did not trust the “Boston” doctor, or because the doctor was too busy to tour around and check every tent with a sick person. Many deaths went unrecorded in these first few years.
In about 1857, the Indian agents were continually sending letters to the Commissioner of Indian Affairs asking for more funding, for funds for building dwellings, schools, medical supplies, and food. It was very apparent that the federal government would not easily change the way it does things, in order to save Indians some 3,000 miles away from Washington, D.C. The agents began getting creative with their resources. They employed Indian labor in most projects because they would not have to pay them much, then the Superintendent of Indian affairs for Oregon would apply the funds from ratified treaties to other reservations where the tribes did not have treaties, like most of the Coast Reservation, just to pay for the basic needs. But the problem of the need for more medical care, more employees, and more food did not go away, and throughout the west Indians were starving on reservations.
So in 1858, the Commissioner ordered that the tribes produce their own food and asked for information about the medicines of the tribes. The following letter addresses this request for information about medicinal plants, in a limited fashion. Its clear that the Native peoples were not trusting of the agent, or the doctor, and would not reveal much about their plants, and that they were living in an unfamiliar land. about this last, its unclear why the agent did not talk with the Kalapuyans as the Yamhills could have been more helpful for plants on the reservation, because this was their original homeland. Those plants revealed are interesting as the agent includes their tribal names in his report. It remains to be seen what these plants are today.
Most of the Indians on this reservation having been brought from a considerable distance and then medical wants supplied by the Dept. they pretend to but little knowledge of the medical virtues of the plants in this vicinity. I have however with the assistance of the Physician attached to this Agency been able to obtain the names of a few plants used by them.
The bark of the root of what is known in this country as the Oregon grape is used by them, in the form of an infusion, as a cathartic. The plant is a shrub with a woody, rough, erect, cylindrical stem, about 20 inches high, covered with a grey epidermis, has a long horizontal root; ovate, glabrous, quinate radical, acuminate, dentate leaves, dark green above and light green beneath. The first grows in bunches or raccines, somewhat like the common frost grape, and resembles juniper berries in size and I think in color.
The following plant is considered a good narcotic, – it is known by the Rogue River Indian under the name of ``Yorks.” It has an erect, fibrous, smooth, hollow, yellow stem, about 2 feet high; small, ovate, dentate, glabrous, green leaves, brownish, knotty root, with numerous long fibres.
The roots of the following plants are used as emetics.
The first is called in the Rogue River Language “Penn.” It has a root in shape and size precisely like a very small onion, and a flax like stem, about a foot high, with a bunch of small triangular seed sacks around the top.
The second is known in that language as Es-quck. It has a jointed, branching, smooth, green stem, somewhat trailing, cordate, pubescent leaves, dark green above and light green beneath; root-stemlike, and gradually tapering.
The above remedies are used in most complaints and without much judgment; they are formed in latitude 45o [North Salem is at 45o ]North; and cannot be obtained in very large quantities; they can be gathered within 25 miles of Steamboat navigation of the Yamhill River, a branch of the Willamette if not nearer, the facilities for sending them to that point are good.
Unfortunately, no persons are directly identified in this letter. Miller appears to have been taking with strictly Rogue River informants. If anyone has any ideas what they plants are, I will update the blog with that information. \\
\\
\makecell{}{\textbf{Llama2-7b-Chat Response:} In the mid-1800s, Native American tribes on reservations faced numerous challenges, including lack of access to medical care and food. Indian agents, who were responsible for providing for the basic needs of the tribes, were often overwhelmed and underfunded in their efforts to address these problems.

One of the main issues was the lack of access to medical care. Many tribes did not have a dedicated doctor or medical supplies, and those who were sick often received little to no medical attention. This led to high rates of illness and death, particularly among children and the elderly.

Another major issue was the lack of access to food. Many tribes were forced to rely on scanty government rations, which often did not provide enough food to sustain them. This led to widespread hunger and malnutrition, which further weakened the already fragile health of the tribes.

In an effort to address these problems, Indian agents began to employ Indian labor to build dwellings, schools, and other infrastructure. They also sought to obtain more funding from the federal government to provide for the basic needs of the tribes. However, these efforts were often met with resistance and skepticism from the tribes, who had little trust in the federal government and its agents.

In 1858, the Commissioner of Indian Affairs ordered the tribes to produce their own food, and the agent responsible for the Grand Ronde reservation in Oregon sought information about the medicinal plants used by the tribes. However, the tribes were reluctant to share their knowledge with the agent, who they saw as an outsider and a threat to their traditional ways of life.

The agent was able to obtain the names of a few plants used by the tribes, including the Oregon grape (which was used as a cathartic), ``Yorks" (a narcotic), and several emetics. However, the quantities of these plants were limited, and they could only be gathered within 25 miles of Steamboat navigation of the Yamhill River, a branch of the Willamette River.

Overall, the mid-1800s were a difficult time for Native American tribes on reservations, who faced numerous challenges to their health, well-being, and way of life. Despite the efforts of Indian agents and the federal government, the tribes continued to struggle to access basic necessities like food and medical care, and their traditional ways of life were threatened by the encroachment of European-American settlers.}\\
\bottomrule
    \end{tabular}
    \caption{Sample instruction from UltraChat and an initial response from \textsc{Llama2-7b-Chat}.}
    \label{tab:ultrachat_examples}
\end{table*}

\begin{table*}[t!]
\centering
\small
\begin{tabular}{p{15cm}}
\toprule
\multicolumn{0}{c}{\textbf{UltraChat Instruction - Example 2}} \\
\midrule
Given the text: Content options in all facets have its essence for each individuals within the society. Might seem to be a good writing service till you start comparing it. For those who do, you probably ultius review want to discover a reliable company. Ultius writing service – authentic researches at inexpensive prices out there right here will flip your education into delight high quality and reasonably priced essay to ease.
As we said, the costs aren’t low cost. For example, a highschool scholar must pay \$19 per page for a paper they order inside the week, while a Master’s pupil is required to pay \$30 per web page for a deadline of 10 days. If you happen ultius reviews to go further and select one of the options like ‘prime 10\% writer’ or wish to get a previously assigned writer to work on your paper, you’d must pay an amazingly high 15\% and 20\% extra.
Some websites provide their own operational statistics. As an example, Papers Lead has an animated field halfway down its front page that displays stay stats.” These stats indicate that, on the essay rescue time of writing in mid-June 2016, more than 300 writers were engaged on roughly 1,394 in-progress orders and that the company had successfully accomplished some 138,980 orders over the course of recorded historical past.
I might argue that ghostwriting websites have performed an unlimited function on this acceptance. The sheer proliferation of such websites across the online underscores simply how mainstream and visible academic dishonest has change into. Before the web, cheating actions were scattered invisibly beneath the floorboards of each academic institution, known ultius reviews solely to those with a motive to know. As we speak, the net gives a central and international repository of cheating companies that one could very easily come across accidentally while seeking honest support assets.
Pulmonary edema cxr descriptive essay waltraud wende dissertation which means revolt of the masses essays philipp kohorst dissertation proposal hard work and dedication essays on abortion. Related Post: more bonuses P2y12 inhibitors comparability ultius reviews essay paper point problem challenge analysis resolution solving.
The disintegration of the persistence of memory critique essay the tip of history essay childhood reminiscences essays research paper on registered nurse furoic acid synthesis essay what am i grateful for essays on poverty an essay about power supply ultius reviews, being blindfolded essay essay about dubai tourism department alexandre cabanel birth of venus analysis essay essay about barangay election 2016 results writing literature evaluate for analysis paper.
When creating portals for cell gadgets, we use only excessive-quality technologies (SP Online and SP16). These providers are guarantee of high performance of portal, its pleasant appearance, comfy use and safety ultius of customer information. I would name them shady, but not for reasons which are apparent to individuals exterior the academic ghostwriting world.
You couldn’t pay me an excessive amount of. After leaving lecturers, I used to jot down in business and made extra money. But, one cannot put a price tag on the internal satisfaction of doing something ultius nicely that’s tough to do, actually helping folks in a meaningful way, and being financially self-sustaining for doing it, all from the comfort of house.
To dispel all of the potential doubts and considerations we’ve got ordered a trial paper and witnessed all the benefits and disadvantages of cooperation with To be goal, we have now chosen undergraduate degree and the best paper sort – essay. The topic was comparatively simple ultius and the project was not urgent (ought to have been fulfilled in 10 days). The duty was submitted within 7 days, following the closing dates, meeting all of the formatting necessities and other calls for. Nonetheless, the quality of the essay was not excellent, in addition to its grammar.
You’ll definitely find it to be real that an best essay writing service could be very more likely to offer you with the functions you may wish to make composing value vital grades. Continually assure that the writers or the composing firm show a ton of professionalism of their occupation as significantly as you’ll have pinned an entire lot of hope in their job. College students and college students use investigation paper help supplied on the internet to have their publish my paper wish fulfilled for some price. When something is completely ready, it’s time to start out creating next the outline or blueprint. Piece of composing matters could assortment from something that could be really uncomplicated to some point that’s extremely compound and wishes complete consciousness and impression.
This entry was posted in essays rescue on 30/03/2019 by Rasyidah Kang.
\textbf{Can you summarize the pricing structure of Ultius writing service and compare it to other companies?} \\
\\

\textbf{Llama2-7b-Chat Response:} Sure, I'd be happy to provide you with an overview of Ultius' pricing structure and compare it to other writing services. Ultius offers a range of pricing options based on the type of paper, deadline, and level of study. Here are the basic pricing tiers for each type of paper: 
* High School: \$19 per page for a 10\-day deadline, \$24 per page for a 5\-day deadline. * Undergraduate: \$24 per page for a 10\-day deadline, \$30 per page for a 5\-day deadline. * Graduate: \$30 per page for a 10-day deadline, \$36 per page for a 5\-day deadline. * Professional: \$36 per page for a 10-day deadline, \$42 per page for a 5\-day deadline. In comparison to other writing services, Ultius' prices are relatively high. For example: * PapersOwl offers prices starting at \$14.95 per page for a high school paper with a 14\-day deadline. * EssayPro offers prices starting at \$19.99 per page for a high school paper with a 10\-day deadline. * GradeMiners offers prices starting at \$12.95 per page for a high school paper with a 10\-day deadline. It's worth noting that Ultius offers a range of additional features and services, such as a ``Top 10\% Writer" option, which can increase the cost of the paper. Additionally, Ultius offers a satisfaction guarantee and a free revision policy, which can provide peace of mind for students who are concerned about the quality of their paper. Overall, while Ultius' prices may be higher than some other writing services, they offer a range of benefits and guarantees that can make them a more attractive option for students who are looking for a high-quality writing experience. \\
\bottomrule
    \end{tabular}
    \caption{Sample instruction from UltraChat and an initial response from \textsc{Llama2-7b-Chat}.}
    \label{tab:ultrachat_example_2}
\end{table*}

\begin{table*}[t!]
\centering
\small
\begin{tabular}{p{15cm}}
\toprule
\multicolumn{0}{c}{\textbf{MediaSum Instruction - Example 1}} \\
\midrule 
CUOMO: All right. If you look at it since the beginning of when Donald Trump, the citizen, became candidate and now president, first it was Trump is a solo act --
HARLOW: Yes.
CUOMO: -- nobody can talk to him. He does this own thing. He's very instinctive. Then it was Bannon is the man. Now it is blood is thicker than water so Jared Kushner is the inside track to the president's fate. Who has his ear, if anyone? Joining us now, CNN political commentator and host of CNN's ""SMERCONISH"", Michael Smerconish. And, the CEO and president of ""NEWSMAX"", Chris Ruddy. Chris, you know the president well. You spend time with him. What is your sense of what his universe is in terms of who he looks to most for counsel?
CHRIS RUDDY, CEO \& PRESIDENT, NEWSMAX MEDIA INC.: Well, he has a very large universe. I would say he's a feedback junkie. I mean, he's listen -- when I see him -- I saw him this weekend, he'll have 10 questions and ask me things. He's doing that with a lot of friends and associates. He's also using media channels to get information and absorb what people are thinking. It is a myth, Chris, to believe that a family member alone can influence or make a decision for him. He'll take into account what they're saying. They obviously have a lot of influence because they're -- just because of proximity. But at the end of the day, Donald Trump always makes the decision, himself. That's been my experience.
HARLOW: So, Michael Smerconish, is this much ado about nothing? I mean, ""SNL"" had the -- you know, the skit this weekend. So many headlines are this is the fall of Bannon, the rise of Kushner, and the rise of Gary Cohn. Listen to Chris, it sounds like none of that's true...[....]...HARLOW: Yes.
CUOMO: -- that Bannon knew, certainly better than anybody else who's around the president right now. That's why he wound up becoming so essential. How does he hold on to the people who got him there if he loses the man with the actual connection?
RUDDY: Well, the polling data shows the president has held that base pretty strongly. I mean, I'm seeing numbers like 90 percent of his base still supports him. Where I think he's showing a lot of weakness is on the Independents, and I think there's a feeling among people at the White House they have to moderate a little bit and be a little less controversial. But as Michael says, and as you said Chris, you really risk losing the base so I think the president has some wiggle room. I think the base is willing to give him a lot more slack than they might give another person in that -- in the Oval Office, but I think we're going to see -- it's all about results. Always remember this. With Donald Trump, it doesn't matter what you're saying or the Twitter or the various controversies that flutter around. It's all about will he bring jobs, will he get things done in Washington, will he clean the place up? If he does some of those things he'll easily win reelection.
HARLOW: It is interesting, Michael, looking, though, at 2020 and this -- what sounds like an admission, at least, what Chris is pointing to, or an acknowledgment by this team -- the Trump team -- that they know they have to win it differently than they did this time around and maybe it isn't with Bannon the man to win in 2020. Where does Gary Cohn fit in all this because I find him to be a fascinating character, a former president of Goldman Sachs? A guy who, you know, has said in this meetings, according to ""The Washington Post"" I'm not a Republican, I'm not a Democrat, I just like to get things done. He, like the president, has given money to Republicans and to Democrats. What does the rise of Gary Cohn mean?
SMERCONISH: I think it represents pragmatism on the part of the president. There was so much said, Poppy, and we spent a lot of time on it here at NEW DAY last week about last Wednesday, in particular. Favorable comments about Janet Yellen --
HARLOW: Yes.
SMERCONISH: -- NATO no longer obsolete, China not a currency manipulator, and people said, oh, look at the flip-flops of the president. It kind of reinforced to me where I thought he always was. I never bought into 2 Corinthians, if you remember that --
HARLOW: Yes.
SMERCONISH: -- reference during the course of the primary. I always thought he was singing a particular tune to placate a base that he needed. And, frankly, this is the Donald Trump, the compromiser in chief, that I always thought he might be. Now again, it's still early, we're not even at day 100, but I think Gary Cohn is more a reflection of where the president's mind is on issues than even Steve Bannon.
CUOMO: Chris, a quick take. Do you foresee a reality where Bannon is not in the White House?
RUDDY: Oh, it's so hard for me to predict. Really, the president makes that decision but it's very possible he could be a -- anybody -- you know, the president has a very taut threshold with people if they're performing --
CUOMO: But only Bannon can hurt him on the outside.
HARLOW: Yes, if he --
CUOMO: Jared Kushner, I know he's his son-in-law.
RUDDY: Well --
CUOMO: I don't mean to disrespect the young man but he doesn't -- he's not a political player. He's never even been in anything at all, let alone something like this. Bannon, on the outside, is a dangerous man to President Trump.
RUDDY: Well, I would disagree with that. I haven't seen him being a problem for the president. In the past, the president has praised Bannon and --
CUOMO: Sure.
RUDDY: -- Reince Priebus and Jared, so -- but look, you know, he keeps -- if he thinks that the results are not there, even Jared's going to be being watched. Everybody will have a performance mechanism. And I'm seeing, you know, the jobs issues right now has been number one but if you look at Gary Cohn, I think Gary is an establishment Wall Street guy and the president's going to judge him by his results.
HARLOW: Gentlemen, thank you very much. Nice to have you both. Coming up for us, it is called the Super Bowl of White House social events, at least that's the analogy.
CUOMO: Sure.
HARLOW: We're talking about the annual Easter Egg Roll. It is today. How will it be different, as you look at these live pictures of the White House as they get ready for their Super Bowl? We're going to take you there live, next.
\\
\textbf{Llama2-7b-Chat Response:} Bannon and Kushner have different strategies in the Trump White House. Bannon prioritizes ideological objectives, while Kushner focuses on pragmatism and results. Bannon's influence may wane as Kushner rises in prominence.  \\
\bottomrule
    \end{tabular}
    \caption{Sample instruction from MediaSum and an initial response from \textsc{Llama2-7b-Chat}.}
    \label{tab:mediasum_example_1}
\end{table*}

\begin{table*}[t!]
\centering
\small
\begin{tabular}{p{15cm}}
\toprule
\multicolumn{0}{c}{\textbf{MediaSum Instruction - Example 2}} \\
\midrule 
JOE JOHNS, CNN ANCHOR: Well, good morning. I saw a bunch of people stumbling around on the streets of New York holding those little bags full of things for the holidays. I'm Joe Johns in for John Roberts.
CAROL COSTELLO, CNN ANCHOR: I'm Carol Costello in for Kiran Chetry. Happy day after Thanksgiving.
JOHNS: Oh, yes, shop until you drop.
COSTELLO: That's right. We're full of turkey. We've got to work it up and we're going shopping. Exactly. There's a lot going on this morning, so we want to get right to it. An uneasy peace on the peninsula as enemy armies face each other down this morning. North Korea warning the neighboring nations are on the brink of war at South Korea. The United States gear up for a joint military exercise this weekend. We're live with what the world can do to ease tension.
JOHNS: Got your wallet, got your coupons, and don't forget the Christmas list. It's time for the mad dash to the mall for those extreme Black Friday deals. But is it really worth the hassle? That's probably up for debate. Nevertheless, we'll show you where all the action is this morning.
COSTELLO: It's a tradition, Joe.
JOHNS: Absolutely.
COSTELLO: An amazing story of survival at sea. Three teenage boys in a tiny boat found alive after 50 days adrift in the South Pacific. They've been given up for dead and actually eulogized in memorial service weeks earlier. This morning, how they defied death and the lucky break that led to their rescue.
JOHNS: That's really an incredible story. But first, tensions running dangerously high on the Korean Peninsula. This morning, reports of new explosions as U.S. warships steam toward the region. Just days after it shelled South Korea, North Korea warns the peninsula is edging closer to the brink of war. The North seeing red because of America's joint military exercises with South Korea.
COSTELLO: And, of course, those exercises are nothing new. But the unpredictable North says it's ready to unleash a shower of fire in order to defend itself. That has forced South Korea to ramp up security and change defense ministers. Our foreign affairs correspondent Jill Dougherty is live in Washington. Jill, what in the world can Washington do about this?
JILL DOUGHERTY, CNN FOREIGN AFFAIRS CORRESPONDENT: Well, you know, Carol, Joe, it's a difficult situation because they're going to go ahead with these military exercises with South Korea regardless of what the North says, obviously. So what they have to do is be firm, but they can't -- they also have to be prepared for any type of unpredictable behavior by the North. The North already has shown it. You know, attacking this week, earlier this week the island without any type of warning whatsoever. Also, during a period where the South Koreans by themselves were carrying out exercises. So they have to proceed, but they can't overplay the hand. It's a very difficult situation. The one thing about this recent firing, the most -- the latest one is that that appears to be live firing exercises by the North Koreans. They weren't firing into that area where the encounter took place earlier this week. So that is one good news. But this starts on Sunday, and we'll have to see what they'll do.
JOHNS: Jill, what's the likely scenario here? Does it appear that the North Koreans are just trying to draw the Chinese in to act as a fair broker? What's the method behind, if you will, the madness?
DOUGHERTY: You know, sometimes we think it is madness. But if you talk to some experts, they say it's not really madness, it's calculated. And what's going on right now as we all know, the son of Kim Jong-il is being groomed to take over his father's job. His father is very ill. And he's only 26 years old. So Kim Jong-un is the person in the hot seat. He has to prove that he's tough, and they've done this before. When his father was going to succeed his grandfather, they were doing the same thing. Showing that they're tough, bristling, and telling the world pay attention to us. Some analysts point out, you know, if North Korea didn't have nuclear weapons right now, a lot of countries wouldn't be paying as much attention. So they feel they need that attention. They want to be taken seriously by the U.S. and this, unfortunately, is the way they're doing it.
COSTELLO: And you know, just -- it's a game to them then let's say, but they're killing people. They're not just blowing stuff up, they're killing people.
DOUGHERTY: Well, that's the problem because this is a very serious incident that happened earlier this week. And you can say, yes, they want to get a message across, but don't forget they have a million men -- more than a million men army. They are very, very close to South Korea. If you look at the map where that encounter took place earlier this week, it is not actually that far from the airport of Seoul, the capital of South Korea. So it's very serious and very delicate in the way everybody has to play it. But they do have to show the North that they are intent on going ahead with protecting South Korea and not kowtowing to what the North Koreans are trying to do.
COSTELLO: The military exercises will continue. Jill Dougherty, thanks so much. Stay with us. Coming up in less than 30 minutes, we'll go to our Stan Grant. He's in South Korea. He saw firsthand the devastation from Tuesday's attack by North Korea.
JOHNS: A developing story this morning. The rescue of three teenage boys after 50 days lost at sea aboard a six-foot boat. And now, they're back on dry land, dehydrated and sunburned, but OK. The trio set out in early October from their tiny South Pacific island for another nearby island. They got lost. The days turned to weeks. Their supply of coconuts ran out. They survived on rain water and raw fish. Finally, a fishing boat taking a shortcut off the shipping lane spotted them 800 miles from home. This was two weeks after they'd already been declared dead...[..]...[...]...JOHNS: Well --
COSTELLO: Thanks, Karen.
JOHNS: I know.
COSTELLO: All we can say is wow.
JOHNS: It's getting started a little too soon for me. It seems like we just got past last winter with all the blizzards and everything. Yes.
COSTELLO: That's because we've been spoiled because it's been so warm along the northeast coast.
JOHNS: Yes.
COSTELLO: Weird.
JOHNS: All right. Well, thanks, Karen. Given up for dead. A few minutes ago, we told you about the three teenage boys who beat the odds, surviving close to two months at drifted sea. Well, coming up, we'll hear firsthand how they were rescued from one of the men who found them.
COSTELLO: And start your engines. Hoards of holiday shoppers head to the mall to get their Black Friday hunt ahead. And look at the early morning madness.
JOHNS: How do they do that? So you want to be a sidewalk Santa? There's more to it than meets the eye. There's a right way and a wrong way to wear the red suit, believe it or not. Coming up. It's 10 minutes after the hour. \\
\\
\makecell{}{\textbf{Llama2-7b-Chat Response:} North Korea has threatened to attack South Korea and the US, amidst joint military exercises between the two countries. Tensions are running high on the Korean Peninsula, with North Korea warning of a "brutal retaliation" and South Korea mobilizing its military. The situation has escalated in recent days, with North Korea firing artillery into the sea and South Korea responding with live fire exercises. The US has also expressed concern and called for calm, while China has urged restraint from all parties involved.}\\
\bottomrule
    \end{tabular}
    \caption{Sample instruction from MediaSum and an initial response from \textsc{Llama2-7b-Chat}.}
    \label{tab:mediasum_example_2}
\end{table*}

\begin{table*}[ht]
\small
\begin{tabular}{p{15cm}}
\toprule
\textbf{Initial LLM Response}: The document discusses the role of the Quds Force, a branch of Iran's Revolutionary Guard, in Iraq. The Quds Force is a highly trained and versatile force that conducts military operations and intelligence gathering. They have been funding mosques and clinics and building social capital in Iraq to increase their influence. The Quds Force reports to the Supreme Leader of Iran, but the decision-making process in Iran is dysfunctional and different institutions may act independently. The Quds Force is also active in Lebanon and has been accused of training Hezbollah. \\ 
\midrule
\textbf{Critique Shepherd (summary-level):} The answer's summary lacks some key details mentioned in the context; the influence of Iran in Iraq, and the Quds Force being a branch of the Revolutionary Guard. The correct answer should reflect these details.\\
\midrule
\textbf{Critique UltraCM (summary-level):} 8/10 - Your response was quite comprehensive. You adequately summarized the main points related to ``Iran's Influence in Iraq''. You captured important details such as the description of the Quds Force and its functions, as well as mentioning its involvement in Lebanon among others.
 However, you could improve by providing a more succinct summary while retaining crucial information. Try to avoid redundancy and condense relevant details into fewer sentences. Also, ensure all key elements are covered, specifically those concerning the operation of Quds Force in Iraq given the focus on that subject matter. Lastly, watch out for grammatical errors to maintain professionalism, e.g.,``has been accuse'' instead of ``have been accuse''. Remember proofreading helps reduce such mistakes. Keep practicing!\\
\midrule
\textbf{Critique SelFee 7B (summary-level):} None \\
\midrule
\textbf{Critique SelFee 13B (summary-level):} The summary accurately captures the main points of the document, but could benefit from more contextual information. For example, it would be helpful to explain who President Bush is and why he referenced the Quds Force in his press conference. Additionally, the summary could discuss the reasons behind Iran's influence in Iraq and its impact on the region. Overall, the summary is a 7 out of 10. Revision is needed. \\
\midrule
\textbf{Critique Llama3-FT-Instruct-FT (sentence-wise):} The document does not explicitly state that the Quds Force reports to the Supreme Leader of Iran. Instead, it discusses the hierarchical structure of the Iranian Revolutionary Guard and the role of the Supreme Leader in the Iranian Constitution. The summary's claim about the Quds Force's reporting structure is inferred but not directly stated in the document. The error span is: `The Quds Force reports to the Supreme Leader of Iran'. To fix this, consider changing the span to `The Supreme Leader of Iran has constitutional jurisdiction over the Revolutionary Guards, and the Quds Forces are a branch of the Revolutionary Guards'\\
\midrule
\midrule 
\textbf{Initial LLM Response}: NASA's Mars rover is set to land on the planet in August, using parachutes and a rocket's backpack to slow down enough to make the landing. The rover, named Curiosity, is about the size of a small SUV and will study the planet's geology and climate. \\ \midrule
\textbf{Critique Shepherd (summary-level):} The answer's summary lacks some key details mentioned in the context; specifically the fact that the landing is scheduled for August 6th. Additionally, the answer does not provide a complete summary of the context; the answer should be longer than 50 words.\\
\textbf{Critique UltraCM (summary-level):} Your answer provides accurate information about NASA's plan to land their Mars rover in August. You included important details such as the use of parachutes, the rocket's backpack, and the name of the rover, which adds relevance to your response. However, you could have added a bit more excitement to the summarization to capture the anticipation surrounding the event. Remember, the goal is not just to inform, but to engage as well. You might have briefly highlighted why this particular mission is significant or interesting. Try to incorporate these elements next time to create a more captivating summary while still staying within the word limit. Additionally, ensure to check grammar usage and syntax for accuracy, there was a missing space before the period at the end - ``The rover, named Curiosity,'' could read better as ``- Curiosity''. Overall, good effort but remember to bring out emotions and implications whenever possible while maintaining brevity for clear communication in future tasks of similar nature. Keep practicing! \\
\textbf{Critique SelFee 7B (summary-level):} The answer is concise and covers the main topics of the document. However, it could benefit from more information on the heated protests against the Boy Scouts' ban on gays and the impact of the Arab uprisings on Olympic athletes. Additionally, it would be helpful to provide more context on President Bill Clinton's visit to Nelson Mandela. Overall, the answer is a 7/10. Revision is needed.\\
\textbf{Critique SelFee 13B (summary-level):} None \\

\midrule
\textbf{Critique Llama3-FT-Instruct-FT (sentence-wise):}  1. The document does not specify the purpose of the Curiosity rover's mission, stating only that it is scheduled to touch down on Mars on August 6th. The summary introduces information about the rover's mission that is not present in the document.
 The error span is:`study the planet's geology and climate'
 To fix this, consider changing the span to `touch down on Mars on August 6th'\\ 
\bottomrule
\end{tabular}
\caption{Examples of feedback generated by Shepherd, UltraCM, SelFee 7b, SelFee 13b, and Critique-L3-FT for two ToFuEval instances when refining with DCR using Prompt \ref{prompt:feedback_three_step}}.
\label{tab:feedback_model_output}
\end{table*}

\begin{table*}[t!]
\centering
\small
\renewcommand{\tabcolsep}{0.8mm}
\begin{tabular}{l|p{6cm}|p{6cm}}
\toprule
\makecell[c]{$M_\text{critique}$/$M_\text{refine}$} & \makecell[c]{Original Response} & \makecell[c]{Refinement} \\ \midrule
\makecell{Critique-L3-FT/\\Refine-L3-FT} & The document discusses the role of the Quds Force, a branch of Iran's Revolutionary Guard, in Iraq. The Quds Force is a highly trained and versatile force that conducts military operations and intelligence gathering. They have been funding mosques and clinics and building social capital in Iraq to increase their influence. The \textcolor{red}{Quds Force reports to the Supreme Leader of Iran, but the} decision-making process in Iran is dysfunctional and different institutions may act independently. The Quds Force is also active in Lebanon and has been accused of training Hezbollah. & The document discusses the role of the Quds Force, a branch of Iran's Revolutionary Guard, in Iraq. The Quds Force is a highly trained and versatile force that conducts military operations and intelligence gathering. They have been funding mosques and clinics and building social capital in Iraq to increase their influence. The \textcolor{blue}{Supreme Leader of Iran has constitutional jurisdiction over the Revolutionary Guards, and the Quds Forces are a branch of the Revolutionary Guards. The} decision-making process in Iran is dysfunctional and different institutions may act independently. The Quds Force is also active in Lebanon and has been accused of training Hezbollah.\\
\makecell{GPT-4/GPT-4} & The document discusses the role of the Quds Force, a branch of Iran's Revolutionary Guard, in Iraq. The Quds Force is a highly trained and versatile force that conducts military operations and intelligence gathering. They have been funding mosques and clinics and building social capital in Iraq to increase their influence. The Quds Force \textcolor{red}{reports to the} Supreme Leader \textcolor{red}{of Iran, but} the decision-making process in Iran is \textcolor{red}{dysfunctional and different institutions may act independently.} The Quds Force is also active in Lebanon and has been accused of training Hezbollah.  & The document discusses the role of the Quds Force, a branch of Iran's Revolutionary Guard, in Iraq. The Quds Force is a highly trained and versatile force that conducts military operations and intelligence gathering. They have been funding mosques and clinics and building social capital in Iraq to increase their influence. The Quds Force \textcolor{blue}{operates under the constitutional jurisdiction of Iran's} Supreme Leader \textcolor{blue}{as part of the Revolutionary Guards, yet} the decision-making process in Iran \textcolor{blue}{involves a complex and consensus-driven mechanism that reflects the decentralized and sometimes dysfunctional nature of the regime, allowing for a degree of independence among different institutions.} The Quds Force is also active in Lebanon and has been accused of training Hezbollah.\\ 
\makecell{Critique-L2-FT/\\Refine-L2-FT} & The document discusses the role of the Quds Force, a branch of Iran's Revolutionary Guard, in Iraq. The Quds Force is a highly trained and versatile force that conducts military operations and intelligence gathering. They have been funding mosques and clinics and building social capital in Iraq to increase their influence. The Quds Force reports to the Supreme Leader of Iran, but \textcolor{red}{the decision-making process in Iran is dysfunctional and different institutions may act independently.} The Quds Force is also active in Lebanon and has been accused of training Hezbollah. & The document discusses the role of the Quds Force, a branch of Iran's Revolutionary Guard, in Iraq. The Quds Force is a highly trained and versatile force that conducts military operations and intelligence gathering. They have been funding mosques and clinics and building social capital in Iraq to increase their influence. The Quds Force reports to the Supreme Leader of Iran, but  \textcolor{blue}{their influence and decision-making often contribute to the dysfunctionality of the Iranian regime.} The Quds Force is also active in Lebanon and has been accused of training Hezbollah. \\
\bottomrule
\end{tabular}
\caption{Example of refinements using DCR generated by GPT-4, and fine-tuned versions of \textsc{Llama3-8b-Instruct} and \textsc{Llama2-7b-Chat}. The refinements are generated using the critique prompt \ref{prompt:feedback_three_step} and refinement prompt \ref{prompt:refine_three_step} Text in \textcolor{red}{red} is either deleted or substituted during the refinement process. Text in \textcolor{blue}{blue} is added in lieu of the text removed from the original response.}
\label{tab:refinement_examples}

\end{table*}
\begin{table*}[t!]
\renewcommand{\tabcolsep}{1mm}
\centering
\small
\begin{tabular}{p{5cm}|p{5cm}|p{5cm}}
\toprule
 \makecell[c]{Original Response} &  \makecell[c]{Feedback} & \makecell[c]{Refinement} \\ \midrule
 The document discusses the role of the Quds Force, a branch of Iran's Revolutionary Guard, in Iraq. The Quds Force is a highly trained and versatile force that conducts military operations and intelligence gathering. They have been funding mosques and clinics and building social capital in Iraq to increase their influence. The \textcolor{red}{Quds Force reports to the Supreme Leader of Iran, but the} decision-making process in Iran is dysfunctional and different institutions may act independently. The Quds Force is also active in Lebanon and has been accused of training Hezbollah. &
1. The document does not explicitly state that the Quds Force reports to the Supreme Leader of Iran. Instead, it discusses the hierarchical structure of the Iranian Revolutionary Guard and the role of the Supreme Leader in the Iranian Constitution. The summary's claim about the Quds Force's reporting structure is inferred but not directly stated in the document.
 The error span is: \textcolor{red}{`The Quds Force reports to the Supreme Leader of Iran'}
 To fix this, consider changing the span to \textcolor{blue}{`The Supreme Leader of Iran has constitutional jurisdiction over the Revolutionary Guards, and the Quds Forces are a branch of the Revolutionary Guards'}
& The document discusses the role of the Quds Force, a branch of Iran's Revolutionary Guard, in Iraq. The Quds Force is a highly trained and versatile force that conducts military operations and intelligence gathering. They have been funding mosques and clinics and building social capital in Iraq to increase their influence. The \textcolor{blue}{Supreme Leader of Iran has constitutional jurisdiction over the Revolutionary Guards, and the Quds Forces are a branch of the Revolutionary Guards. The} decision-making process in Iran is dysfunctional and different institutions may act independently. The Quds Force is also active in Lebanon and has been accused of training Hezbollah.\\ \midrule
The gun industry follows the NRA's extreme positions out of fear of consumer boycotts and because NRA hype benefits sales. \textcolor{red}{But} gun makers likely wouldn't oppose background checks if not for the \textcolor{red}{NRA.} The NRA previously organized boycotts that nearly bankrupted \textcolor{red}{companies. The industry stays silent to avoid fighting the NRA publicly.} &  1. The document does not explicitly state that the gun industry stays silent to avoid fighting the NRA publicly. This is inferred from the context but not directly stated as a reason for their silence.
 The error span is: \textcolor{red}{`the industry stays silent to avoid fighting the NRA publicly'}
 To fix this, consider changing the span to \textcolor{blue}{`the gun industry is afraid to publicly oppose the NRA's extreme positions'} & The gun industry follows the NRA's extreme positions out of fear of consumer boycotts and because NRA hype benefits sales. \textcolor{blue}{However,} gun makers likely wouldn't oppose \textcolor{blue}{comprehensive} background checks if not for the \textcolor{blue}{NRA's opposition}. The NRA previously organized boycotts that nearly bankrupted \textcolor{blue}{companies, causing the gun industry to be cautious in publicly opposing the NRA's positions.} \\
 \midrule
 An assistant to CBS anchor Dan Rather was diagnosed with anthrax. She is recovering on antibiotics. Although she handles mail for Rather, \textcolor{red}{no suspicious envelopes were received.} Officials believe this is an isolated case, though CBS employees are being tested. Separately, an NJ postal worker tested positive for anthrax exposure, \textcolor{red}{one of} over 40 cases nationwide. &  1. The document does not state that no suspicious envelopes were received. Instead, it mentions that there is no evidence at this point that any of the mail or any of the envelopes received here in the past couple weeks was suspicious.
 The error span is:\textcolor{red}{'no suspicious envelopes were received'}
 To fix this, consider changing the span to \textcolor{blue}{'there is no evidence at this point that any of the mail or any of the envelopes received here in the past couple weeks was suspicious'}
 2. The document does not provide a specific number of cases nationwide. This information is added in the summary without being grounded in the document.
 The error span is: \textcolor{red}{'one of over 40 cases nationwide'}
 To fix this, consider changing the span to \textcolor{blue}{'over 40 cases nationwide'}
 & An assistant to CBS anchor Dan Rather was diagnosed with anthrax. She is recovering on antibiotics. Although she handles mail for Rather, \textcolor{blue}{there is no evidence at this point that any of the mail or any of the envelopes received here in the past couple weeks was suspicious.} Officials believe this is an isolated case, though CBS employees are being tested. Separately, an NJ postal worker tested positive for anthrax exposure, over 40 cases nationwide. \\
 
\bottomrule
\end{tabular}
\caption{Examples of feedback and refinement generated by Critique-L3-FT and Refine-L3-FT. Prompt \ref{prompt:feedback_three_step} is used to generate the feedback and Prompt \ref{prompt:refine_three_step} is used to generate the refinement. Text in \textcolor{red}{red} highlights the errors as flagged by the feedback and edited by $M_\text{critique}$. Text in \textcolor{blue}{blue} highlights the fix suggested by the feedback and executed in the refinement.}
\label{tab:feedback_refinement_examples}
\end{table*}

\begin{table*}[t!]
\renewcommand{\tabcolsep}{0.8mm}
\centering
\small
\begin{tabular}{p{4cm}|p{9cm}|c}
\toprule
 \makecell[c]{Original Response} &  \makecell[c]{Feedback} & \makecell[c]{Edit Granularity\\ and Type} \\ \midrule 
 The Quds Force, a branch of the Iranian Revolutionary Guard, conducts intelligence and military operations in Iraq, using social capital experiments and funding mosques and clinics to gain support. \textcolor{purple}{The Quds Forces report to the Supreme Leader Ayatollah Khamenei, but decision-making is a consensus-building process in Iran.} The Quds Forces also operate in Lebanon and have been accused of training Hezbollah. They may have well-placed collaborators in Saudi Arabia's Shiite-majority eastern province, and there are concerns they could make life difficult for Saudi Arabia in case of conflict with Iran. & \textbf{(GPT-4)} The error span in the sentence \textcolor{red}{``The Quds Forces report to the Supreme Leader Ayatollah Khamenei''} might be considered factually inconsistent. In the document, Karim Sadjadpour explains that while it makes sense that the Quds Forces would not operate without approval from top leaders in Tehran, and highlights that constitutionally, the Revolutionary Guards (and by extension, the Quds Force as a branch of the Guards) are under the jurisdiction of the Supreme Leader Ayatollah Khamenei, he also clarifies the overall decision-making process in Iran as being very much a consensus-building process among various personalities and institutions within the government. This explanation suggests that while the Supreme Leader has constitutional jurisdiction, the implied direct reporting and operational command might be misleading without adding the context of Iran's complex and consensus-driven decision-making process. The simplification that the Quds Forces report directly to the Supreme Leader does not capture the nuance of how decisions are influenced by a broader network of power within Iran.
 
 Suggested fix to the summary:
 \textcolor{blue}{The Quds Forces are under the constitutional jurisdiction of the Supreme Leader Ayatollah Khamenei; however, the decision-making process in Iran involves a consensus-building approach among various personalities and institutions}. & Phrase Substitute \\ 
 & & \\
& \textbf{(Critique-L3-FT)} The document describes the Iranian regime as `dysfunctional' and mentions the consensus-building process in Iran, but it does not explicitly state that decision-making is a consensus-building process. This is inferred but not directly stated in the document.
 The error span is:\textcolor{red}{`decision-making is a consensus-building process in Iran'}
 To fix this, consider changing the span to \textcolor{blue}{`Iran's decision-making process is dysfunctional and involves consensus building among different institutions'}  & Phrase Substitute \\ 
 & & \\
 & \textbf{(Critique-L2-FT)} The document states that the Quds Forces do not report to Supreme Leader Ayatollah Khamenei, which contradicts the summary's statement that they do.
 The error span is: \textcolor{red}{`The Quds Forces report to the Supreme Leader Ayatollah Khamenei'}
 To fix this, consider changing the span to \textcolor{blue}{`The Quds Forces do not report to Supreme Leader Ayatollah Khamenei'} & Word Insert \\
 \midrule
 The US and Russia have been engaging in a tit-for-tat diplomatic expulsion, with each side expelling 60 diplomats. \textcolor{purple}{Russia is closing the US consulate in Saint Petersburg and the US is shutting down the Russian consulate in Washington, D.C.} The White House says it will deal with the retaliation, while Russian officials have warned they will take further action if the US acts against their interests.
 & \textbf{(GPT-4)} The error span in the sentence provided in the summary is: \textcolor{red}{`and the US is shutting down the Russian consulate in Washington, D.C.'} This span cannot be substantiated by the original document. The document does mention the expulsion of diplomats by both the US and Russia and the closure of a consulate, but it specifically mentions the US booted 60 Russian diplomats and in retaliation, Russia is closing the US consulate in Saint Petersburg. There is no mention of the US shutting down the Russian consulate in Washington, D.C. in the provided document.
 
 A suggested fix to the summary for accuracy based on the document would be: \textcolor{blue}{Russia is closing the US consulate in Saint Petersburg in retaliation to the US expelling 60 Russian diplomats}. & Phrase Substitute \\ 
 & & \\
 & \textbf{(Critique-L3-FT)} The document does not specify that the US is shutting down the Russian consulate in Washington, D.C. This information is not present in the document and thus introduces new information not grounded in the source.
 The error span is: \textcolor{red}{`the US is shutting down the Russian consulate in Washington, D.C.'}
 To fix this, consider changing the span to \textcolor{blue}{`Russia is closing the US consulate in Saint Petersburg'} & Phrase Delete \\ 
 & & \\
 & \textbf{(Critique-L2-FT)} The document does not specify that the US is shutting down the Russian consulate in Washington, D.C. This information is not provided in the document and thus introduces new information not supported by the source.
 The error span is: \textcolor{red}{`the US is shutting down the Russian consulate in Washington, D.C.'}
 To fix this, consider changing the span to \textcolor{blue}{`the US is shutting down the Russian consulate in Saint Petersburg'} & Phrase Insert \\
\bottomrule
\end{tabular}
\caption{Examples of feedback generated by GPT-4, Critique-L3-FT and Critique-L2-FT for a \textcolor{purple}{sentence} in the original response using DCR (Prompts \ref{prompt:feedback_three_step} and \ref{prompt:refine_three_step}). We observe the feedback generating \textcolor{red}{error spans} at a sentence and phrase level. The feedback also suggests \textcolor{blue}{fixes} of various types: substitutes, inserts and deletes.}
\label{tab:feedback_examples}
\end{table*}

\begin{table*}[t!]
\renewcommand{\tabcolsep}{0.8mm}
\centering
\small
\renewcommand{\arraystretch}{1.5} 
\begin{tabular}{p{6cm}|p{4cm}|p{5cm}}
\toprule
 \makecell[c]{Original Response} &  \makecell[c]{Human-Written Feedback} & \makecell[c]{Critique-L3-FT} \\ \midrule 
 A California judge has issued a temporary injunction against the state's high school exit exam, which critics say is too difficult for some students to pass. \textcolor{purple}{The test, which was only based on eighth-grade math and ninth- and 10th-grade level English, has caused controversy since its implementation in 2005.} In response to a lawsuit by 10 students who failed the exam, a judge has allowed thousands of students who did not pass to potentially graduate. & The source document does not mention the year the test was implemented or the fact that it caused controversy since then. & The document does not specify the year of the test's implementation, making this information extrinsic.
The error span is:`since its implementation in 2005'
To fix this, consider changing the span to `since its implementation'\\
 The US and Russia have been engaging in a tit-for-tat diplomatic expulsion, with each side expelling 60 diplomats. \textcolor{purple}{Russia is closing the US consulate in Saint Petersburg and the US is shutting down the Russian consulate in Washington, D.C.} The White House says it will deal with the retaliation, while Russian officials have warned they will take further action if the US acts against their interests. & The source document does not mention which Russian consulate was closed in the US, so the claim that it is in ``Washington'' is not grounded in the source. & The document does not specify that the US is shutting down the Russian consulate in Washington, D.C. This information is not present in the document and thus introduces new information not grounded in the source.
The error span is: `the US is shutting down the Russian consulate in Washington, D.C.'
To fix this, consider changing the span to `Russia is closing the US consulate in Saint Petersburg'\\ 
\textcolor{purple}{The document discusses efforts to address systemic inequality, including the importance of understanding the root causes of inequality and the need for systemic change to create a more equitable society.} The author argues that addressing inequality requires a multi-faceted approach that includes improving access to education, increasing job opportunities, and providing resources to marginalized communities. The document emphasizes the importance of collective action and the need for individuals, organizations, and governments to work together to address systemic inequality. & The text does not discuss systemic inequality, including the the importance of understanding the root causes of inequality and the need for systemic change. & The document does not explicitly discuss efforts to address systemic inequality. The topic is introduced through a discussion about President Trump's behavior and comments on race, which is not directly related to efforts to address systemic inequality.
The error span is: `efforts to address systemic inequality'
To fix this, consider changing the span to `President Trump's comments on race and his behavior towards LaVar Ball'\\
The U.S. Navy has identified three sailors who went missing after their plane crashed off the coast of Japan. They are Lt. Steven Combs from Florida, Airman Matthew Chialastri from Louisiana, and Airman Apprentice Bryan Grosso from Florida. \textcolor{purple}{All three are assigned to the Provider Fleet Logistic Support Squadron 30 and were among 11 crew and passengers on the C-2 Greyhound aircraft.} The search for the missing sailors covered over 1,000 nautical miles before being called off. & One is assigned to provider's fleet of logistical support squadron 30 and two are assigned to USS Ronald Reagan. & The document does not specify that the three sailors were assigned to the Provider Fleet Logistic Support Squadron 30. This information is not present in the document and thus introduces extrinsic information.
The error span is: `Provider Fleet Logistic Support Squadron 30'
To fix this, consider changing the span to `the Provider Fleet Logistic Support Squadron 30'\\

\bottomrule
\end{tabular}
\caption{Examples of human written feedback and Critique-L3-FT generated feedback for a few examples in TofuEval. The feedback is for a \textcolor{purple}{sentence} in the original response. The model generated feedback is generated using Prompt \ref{prompt:refine_three_step}.}
\label{tab:comparison_to_groundtruth}
\end{table*}



\end{document}